\documentclass[11pt]{article}

\usepackage[preprint]{acl}

\usepackage{times}
\usepackage{latexsym}

\usepackage[T1]{fontenc}

\usepackage[utf8]{inputenc}

\usepackage{microtype}

\usepackage{inconsolata}

\usepackage{graphicx}

\usepackage[utf8]{inputenc} 
\usepackage[T1]{fontenc}    
\usepackage{hyperref}       
\usepackage{url}            
\usepackage{longtable}
\usepackage{booktabs}       
\usepackage{amsfonts}       
\usepackage{nicefrac}       
\usepackage{microtype}      
\usepackage{xcolor}         
\usepackage{enumitem}

\usepackage{subfig}
\usepackage{xspace}
\usepackage[table]{xcolor}
\usepackage{multirow}
\usepackage{geometry}
\usepackage{adjustbox}
\usepackage{soul}
\usepackage[most]{tcolorbox}

\definecolor{lightgreen}{RGB}{0, 255, 0}
\definecolor{lightred}{RGB}{255, 204, 204}
\definecolor{myblue}{RGB}{30, 144, 255} 
\definecolor{mediumblue}{RGB}{0, 0, 205}
\definecolor{darkorange}{RGB}{255, 140, 0}

\newtcolorbox{custombox}[1]{ 
  title=#1,
  colframe=black,
  colback=gray!10,
  coltitle=white,
  colbacktitle=black,
  fonttitle=\bfseries,
  sharp corners=south,
  boxrule=0.8pt
}

\newcommand{\bbehqa}{\textsc{DisambiguationQA}\xspace}
\newcommand{\tinytruthfulqa}{\textsc{tinyTruthfulQA}\xspace}

\newcommand{\baseline}{\textsc{Baseline}\xspace}
\newcommand{\chainot}{\textsc{CoT}\xspace}
\newcommand{\selfconsistency}{\textsc{SC}\xspace}

\newcommand{\smol}{\textsc{SmolLM2-1.7B}\xspace}
\newcommand{\qwenthree}{\textsc{Qwen2.5-3B}\xspace}
\newcommand{\llama}{\textsc{Llama-3.1-8B}\xspace}
\newcommand{\qwenfourteen}{\textsc{Qwen2.5-14B}\xspace}
\newcommand{\deepseek}{\textsc{DeepSeek-R1-Distill-Llama-8B}\xspace}
\newcommand{\gemini}{\textsc{Gemini-2.0-Flash}\xspace}

\title{Self-Correcting Large Language Models: Generation vs.~Multiple Choice}

\author{Hossein A.~Rahmani\textsuperscript{$\dagger$}, Satyapriya Krishna\textsuperscript{$\ddagger$}\thanks{Work was done prior to joining Amazon.}, Xi Wang\textsuperscript{$\nabla$}, Mohammadmehdi Naghiaei\textsuperscript{$\diamondsuit$}, Emine Yilmaz\textsuperscript{$\dagger$} \\
        \textsuperscript{$\dagger$}University College London, \textsuperscript{$\ddagger$}Amazon AGI, \textsuperscript{$\nabla$}University of Sheffield, \textsuperscript{$\diamondsuit$}University of Southern California \\
        \texttt{\{hossein.rahmani.22, emine.yilmaz\}@ucl.ac.uk, skrishna@g.harvard.edu} \\
        \texttt{xi.wang@sheffield.ac.uk, naghiaei@usc.edu}}

\begin{document}
\maketitle
\begin{abstract}
Large language models have recently demonstrated remarkable abilities to self-correct their responses through iterative refinement, often referred to as self-consistency or self-reflection. However, the dynamics of this self-correction mechanism may differ substantially depending on whether the model is tasked with open-ended text generation or with selecting the most appropriate response from multiple predefined options. In this paper, we conduct a systematic investigation of these two paradigms by comparing performance trends and error-correction behaviors across various natural language understanding and reasoning tasks, covering language models of different scales and families. Our experimental results reveal distinct patterns of improvement and failure modes: 

\textit{While open-ended generation often benefits from the flexibility of re-interpretation and compositional refinement, multiple-choice selection can leverage clearer solution boundaries but may be limited by the provided options}. This contrast also reflects the dual demands faced by emerging agentic LLM applications: effective agents must not only generate and refine open-ended plans or explanations, but also make reliable discrete choices when operating within constrained action spaces. Our findings, therefore, highlight that the design of self-correction mechanisms should take into account the interaction between task structure and output space, with implications for both knowledge-intensive reasoning and decision-oriented applications of LLMs.\footnote{Codes and experiments are available at \url{https://github.com/rahmanidashti/llm-self-correction}}
\end{abstract}

\section{Introduction}
\label{sec:intro}
Recent advances in Large Language Models (LLMs) have illustrated that iterative self-correction, where a model re-examines and revises its output under a self-reflection framework, can lead to significant performance gains across a variety of tasks \cite[\textit{inter alia}]{madaan2023selfrefine,cook2024self-tick,shinn2023reflexion,gou2024critic}. This emergent ability is often attributed to the models' capacity to integrate chain-of-thought reasoning \cite{kamoi2024can,chang2024survey,wei2022chain}, prompting them to refine their own outputs as addressed by a human proofreader or mentor. Regarding performance validation, existing studies on self-correction have generally focused on free-form text generation \cite[\textit{inter alia}]{huang2023cannot-self-correct,madaan2023selfrefine,zelikman2022star,ma2025s2r,kumar2025training,10.5555/3692070.3693094}, such as dialogue response, code optimization, and acronym generation. These tasks align with the strategy of language model optimization in addressing next token prediction. 

However, as LLM applications expand, evaluation restricted to free-form generation offers an incomplete picture. For instance, NVIDIA advocates the deployment of smaller language models in agentic systems for tasks such as API calls and orchestration with external tools, motivated by sustainability and efficiency considerations \cite{belcak2025small}. This highlights the need to examine self-correction beyond open-ended generation. In this study, we categorize natural language modeling tasks into two broad paradigms: \textbf{free-form text generation} and \textbf{multi-choice prediction}. The former treats modeling as unconstrained sequence generation over the full vocabulary, while the latter frames it as classification over a fixed set of candidate answers. These paradigms are complementary: multi-choice tasks test precise discrimination under constraints, whereas free-form tasks assess expressive generation, and together they capture the main modes of LLM use in applications such as question answering, reasoning, and open-ended dialogue.

In this paper, we investigate how self-correction unfolds when comparing open-ended generation against multiple-choice question scenarios. We hypothesize that while open-ended generation may benefit from enhanced flexibility and creativity, it also faces a larger search space and the risk of compounding errors. By contrast, multiple-choice models operate in a constrained space, which can reduce semantic drift yet limit creative corrections. Our study explores how these respective factors interact with iterative refinement, shedding light on whether self-correction aligns more naturally with either unconstrained or constrained output space.

To address these questions, we conduct comprehensive experiments on two distinct datasets that differ in nature, one focusing on knowledge-intensive question answering and the other on reasoning-oriented problems. We perform iterative inference, giving the model multiple opportunities to reevaluate and revise. By comparing error rates, consistency across iterations, and eventual convergence in each paradigm, we expose nuanced trade-offs in how LLMs adapt to different output constraints under a self-correction regime. Our results provide practical insights for the design and deployment of LLM-based systems, highlighting opportunities to harness better or tailor self-correction behaviors for diverse application settings. Furthermore, we discuss how our findings inform the broader research agenda of aligning emergent capabilities in large-scale models with varied real-world task requirements.

\section{Related Works}
\label{sec:relwork}

\paragraph{Iterative Reasoning and Self-correction in LLMs.}
Large language models first showed an emergent ability to reason step-by-step when prompted with \emph{chain-of-thought} (\chainot) examples \citep{wei2022chain}. Shortly after, \citet{wang2022selfconsistency} demonstrated that sampling several independent reasoning traces and selecting the majority answer---dubbed \textit{self-consistency} (\selfconsistency)---boosts accuracy on arithmetic and commonsense tasks. Follow-up studies made the correction loop explicit by asking the model to critique its own draft before rewriting it, leading to sizeable gains in factual QA and code generation \citep{madaan2023selfrefine}. Variants that call external tools such as Python or knowledge bases during the critique stage further reduce hallucinations in open-ended generation \citep{chen2023program,yao2023tree,gou2024critic}. These works collectively suggest that LLMs can act as both solver and reviewer, but they focus almost exclusively on free-form text outputs.

\paragraph{Verification–based Refinement.}
Instead of trusting the model's final token distribution, several papers add lightweight verifiers. \citet{cobbe2021training} attach unit tests to code synthesis; \citet{du2023improvingfact} use factuality checkers for summarization; \citet{pryzant2023deixis} adopt entailment models for reading comprehension. The common pattern is a two-step pipeline where the LLM proposes an answer, then a cheaper or more precise module scores it. Our work keeps the entire loop inside the language model, isolating the effect of output format itself (generation vs.~multiple-choice) from external verification.

\paragraph{Answer Selection and Multiple-Choice Prompting.}
Tasks with a \emph{closed candidate set} (e.g., MMLU \citep{hendrycks2021mmlu}, ARC \citep{clark2018arc}) are typically solved by mapping each option to an independent prompt and picking the highest-logit answer \citep{brown2020gpt3}. Several groups have tried to retrofit iterative reasoning onto this template. \citet{zhu2024mcrepair} prepend a self-explanation, rescore the options with the explanation as additional context, and report modest but consistent gains. \citet{li2024logitlens} show that calibrating logits with contrastive rationales helps low-parameter models, while \citet{pan2023multiple} explore ensembling diverse rationales. Yet a systematic comparison between correction dynamics in \emph{open} versus \emph{closed} output spaces is missing; our study provides that head-to-head analysis.

\paragraph{Bridging the paradigms.}
Contemporary benchmarks increasingly mix free-form and categorical sub-tasks—e.g., TruthfulQA has both short-answer and multiple-choice splits \citep{lin2022truthfulqa}. Deployment settings such as tutoring agents or search assistants likewise alternate between generating explanations and selecting the best passages. Understanding whether self-correction behaves differently under these two regimes is therefore more than a methodological curiosity as it affects prompt engineering, compute budgeting, and safety guard-rail design. By re-implementing the main correction strategies from the literature under a unified experimental budget, we show that the \emph{shape} of the output space itself controls how much an LLM can benefit from extra reflection rounds.

\section{Open-ended Generation {\normalfont vs.} Multiple-Choice Answer Selection}
\label{sec:paradigms}

Large language models are increasingly expected to handle a wide spectrum of downstream tasks, ranging from unconstrained natural language generation, such as open-domain question answering, to highly structured classification problems, like sentiment analysis. Two of the most commonly encountered settings are (i) open-ended generation, where the model must produce a free-form text response, and (ii) multiple-choice answer selection, where it must select a single correct option from a predefined set of choices. While these two paradigms are often operationalized using the same model architecture and weights, they impose fundamentally different constraints on the output space and influence how self-correction unfolds over successive inference steps. This section formalizes these two paradigms, describes how self-correction mechanisms are instantiated within each, and presents qualitative differences that help explain the empirical patterns observed in Section~\ref{sec:results}.

\paragraph{Open-Ended Generation.} 
In the open-ended generation setting, the model is required to produce an output sequence 
\( y^{(0)} = (y^{(0)}_{1}, \ldots, y^{(0)}_{T}) \in \mathcal{V}^{*} \), 
where \( \mathcal{V} \) denotes the vocabulary and \( T \) is the (variable) sequence length. 
The generation is conditioned on an input \( x \), which may correspond to a question, prompt, or instruction, such that the model defines a conditional distribution: 

\( p(y^{(0)} \mid x) = \prod_{t=1}^{T} p(y^{(0)}_{t} \mid y^{(0)}_{<t}, x) \) 

\noindent This formulation captures the standard auto-regressive decoding process for open-ended text generation. The generated sequence may consist of a sentence, paragraph, or longer passage, and there are no explicit structural constraints beyond \textit{syntactic plausibility} and \textit{task relevance}. 

Self-correction in this paradigm typically proceeds by prompting the model to critique its initial output — either via explicit instructions (“identify any flaws”) or implicit prompting strategies (“think step by step”) — followed by a new generation \( y^{(1)} \). This iterative process can be repeated multiple times, resulting in a sequence \( \{y^{(k)}\}_{k=0}^{K} \), where each revised answer aims to improve upon the previous one. A final answer can be selected using majority voting, log-probability re-ranking, or verifier-based scoring. Because generation is unconstrained, each iteration can introduce new content, restructure previous arguments, or expand omitted details. While this offers flexibility and the potential for substantial improvements, it also opens the door to risks such as semantic drift \cite{ji2023towards,ji2023survey}, where the answer becomes misaligned with the original question over time, or hallucinations, where fictitious facts are introduced in an attempt to improve fluency or apparent coherence. These failure modes tend to accumulate if the model ``over-corrects'' by deviating from the initial context \cite{spataru2024know}.

\paragraph{Multiple-Choice Answer Selection.} By contrast, the multi-choice setting restricts the output space to a finite set of candidate answers \( A = \{a_1, a_2, \ldots, a_M\} \). For each question \( x \), the model computes a logit vector \( \ell(x) \in \mathbb{R}^{M} \), from which a softmax distribution is derived, and selects the most probable answer. Self-correction in this paradigm does not involve rewriting text but rather involves revisiting the initial logits after incorporating additional information. One common strategy is to generate a rationale \( r^{(t)} \) for why a particular answer is correct, then concatenate this rationale to the original prompt and recompute the logits to obtain \( \ell^{(t+1)}(x, r^{(t)}) \) \cite{huang2023cannot-self-correct,liu2024large}. Over successive iterations, this allows the model to refine its beliefs based on its own reasoning. However, since the answer set is fixed, the model cannot explore novel hypotheses or restructure the space of answers; instead, it can only shift probability mass among existing options. This bounded nature of the output space makes multiple-choice settings more stable and less prone to semantic drift, but also potentially less effective at recovering from early errors — especially if the correct answer has low initial probability and the generated rationales fail to meaningfully influence the logits.

\paragraph{Qualitative Differences.} The two paradigms, i.e., open-ended generation and multiple-choice selection, exhibit distinct self-correction dynamics due to their differing output constraints. In open-ended generation, performance gains are typically front-loaded, with the most significant improvements occurring in the first few iterations as the model repairs inconsistencies or fills in missing details \cite{cook2024self-tick,huang2023cannot-self-correct,gou2024critic}. However, this flexibility also increases the risk of semantic drift in later rounds \cite{spataru2024know}: if the model's revisions start to go off-topic or introduce inaccuracies, the session can degrade without external intervention. In contrast, multiple-choice tasks show steadier, more incremental improvements, benefiting from the stability of a fixed answer set. They may suffer, however, from \textit{logit inertia} when the correct option is initially underweighted. The model can be difficult to move to a low-probability answer unless a very compelling rationale shifts the balance. Generation tends to be more compute-intensive due to longer outputs per iteration, while multiple-choice achieves better accuracy-to-token efficiency by focusing on short discriminative outputs. Additionally, model scale interacts differently across formats. Larger models can better mitigate drift in generation through coherent reasoning chains, while smaller models perform more reliably in multiple-choice settings due to the structured nature of the output space and the guidance provided by explicit options.

Understanding these qualitative and quantitative differences between the two paradigms is crucial for designing robust systems that use LLMs in iterative inference settings. Depending on the task requirements, whether correctness, stability, creativity, or inference budget is the primary constraint, one or the other format may be more appropriate, and self-correction strategies should be tailored accordingly.

\section{Experimental Setup}
\label{sec:setup}

\paragraph{Problem Statement.}
In this study, we aim to evaluate the dynamics of iterative self-correction under constrained generation and multiple-choice selection across representative tasks.  Let $x \in \mathcal{X}$ denote an input instance (e.g., a question) with ground-truth answer $y^\star$. An LLM parameterised by $\theta$ produces an initial response $y^{(0)}$ whose format depends on the task paradigm. For \textit{open-ended generation}, the model outputs a sequence $y^{(0)} \in V^{\ast}$ with $p_\theta\!\big(y^{(0)} \mid x\big) \;=\; \prod_{t=1}^{T} p_\theta\!\big(y^{(0)}t \mid y^{(0)}{<t}, x\big)$. In contrast, for multiple-choice selection, the model selects $y^{(0)} \in A=\{a_1,\dots,a_M\}$ from logits $\ell(x)\in\mathbb{R}^M$, i.e., $y^{(0)} \;=\; \arg\max_{a_i \in A} \ell_i(x), \qquad$ $\sigma_i^{(0)}(x) \;=\; \frac{e^{\ell_i(x)}}{\sum_{j=1}^{M} e^{\ell_j(x)}}$. By applying iterative self-correct, given history $\mathcal{H}^{(k-1)} = (x, y^{(0)}, \dots, y^{(k-1)})$, the model produces a revision $y^{(k)} \sim p_\theta\!\big(\cdot \mid \mathcal{H}^{(k-1)}\big), \qquad k=1,\dots, K$.

We study the sequence $\mathcal{Y}(x)=\{y^{(k)}\}_{k=0}^{K}$ and aim to maximize task accuracy of the terminal output $y^{(K)}$ over $x \sim \mathcal{D}$. We seek to observe how performance evolves with successive self-correction iterations and how error correction or degradation manifests in each paradigm. To that end, we set up experiments on two distinct question-answering benchmarks and examine multiple LLMs under various prompting strategies.

\begin{figure*}[!tbp]
  \centering
  \subfloat[Baseline]{
    \includegraphics[width=0.3\textwidth]{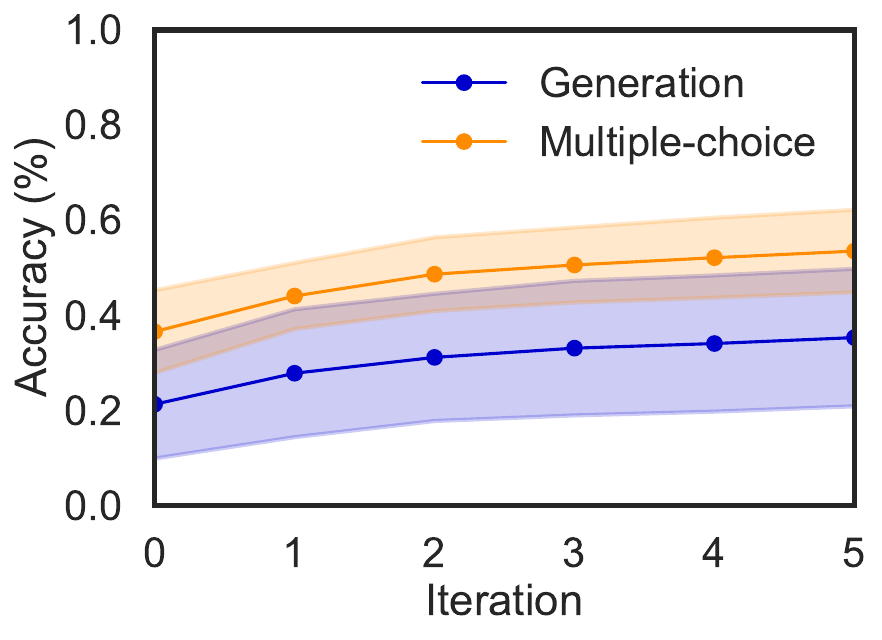}
    \label{fig:f1}
  }
  \subfloat[\chainot]{\includegraphics[width=0.3\textwidth]{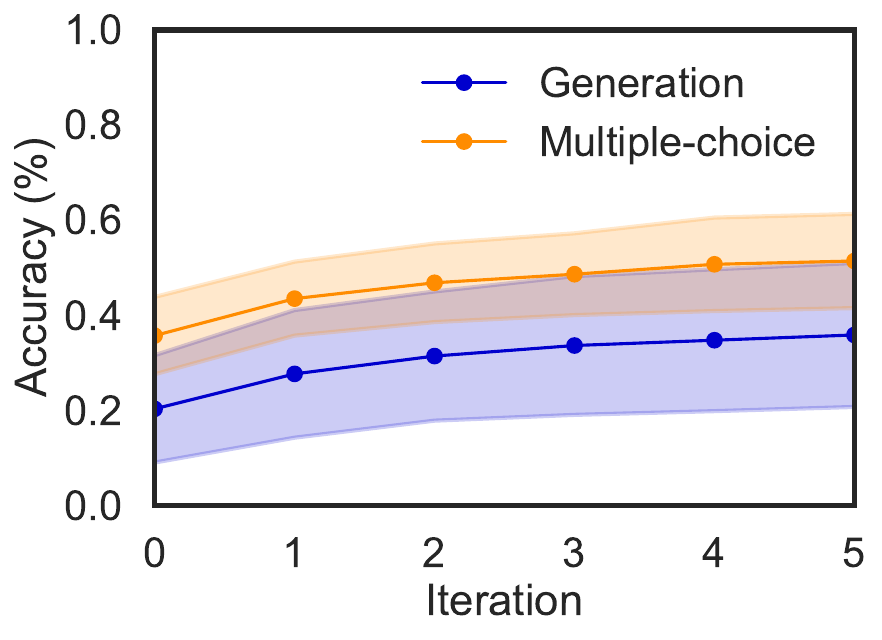}\label{fig:f2}}
  \subfloat[\selfconsistency]{\includegraphics[width=0.3\textwidth]{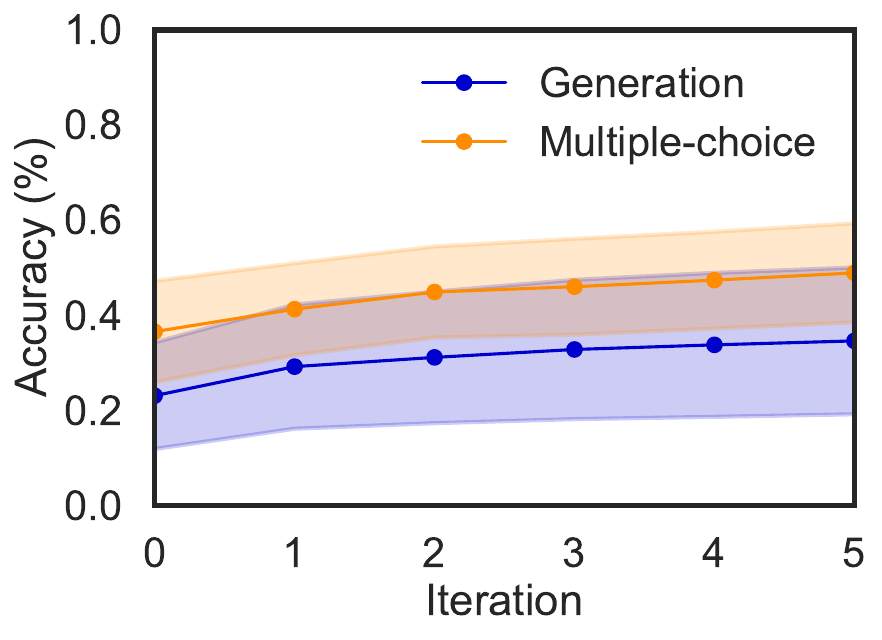}\label{fig:f2}}
  \qquad
  \subfloat[Baseline]{\includegraphics[width=0.3\textwidth]{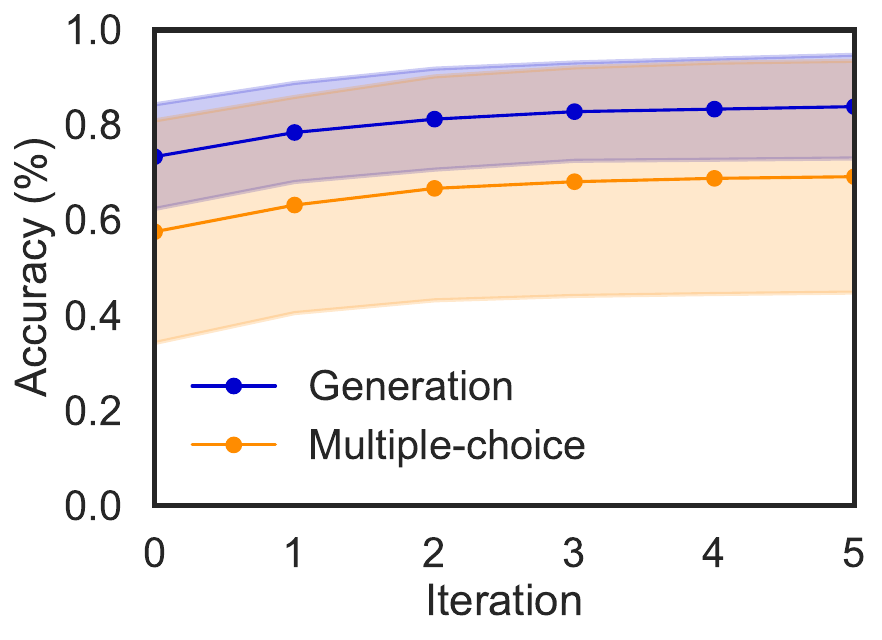}}
  \subfloat[\chainot]{\includegraphics[width=0.3\textwidth]{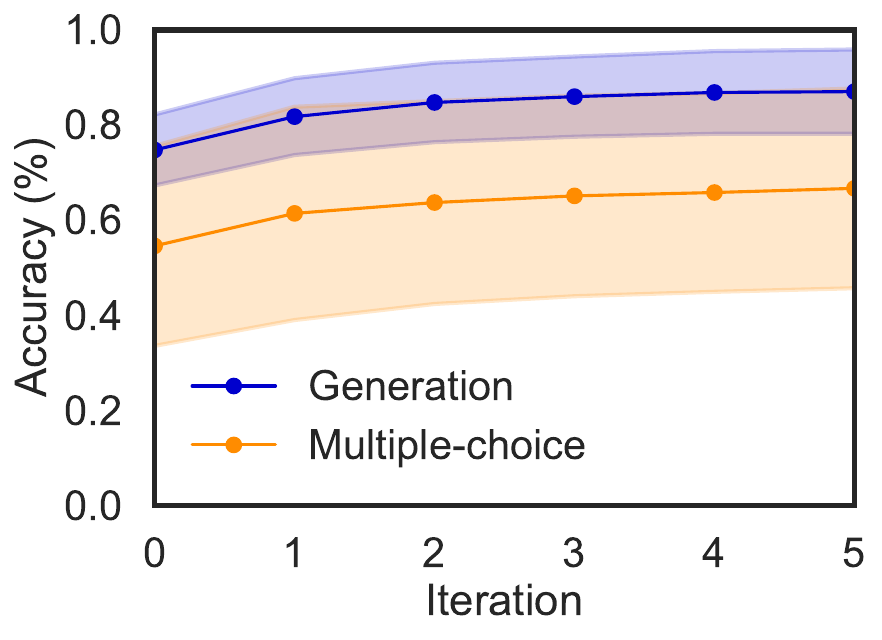}\label{fig:f2}}
  \subfloat[\selfconsistency]{\includegraphics[width=0.3\textwidth]{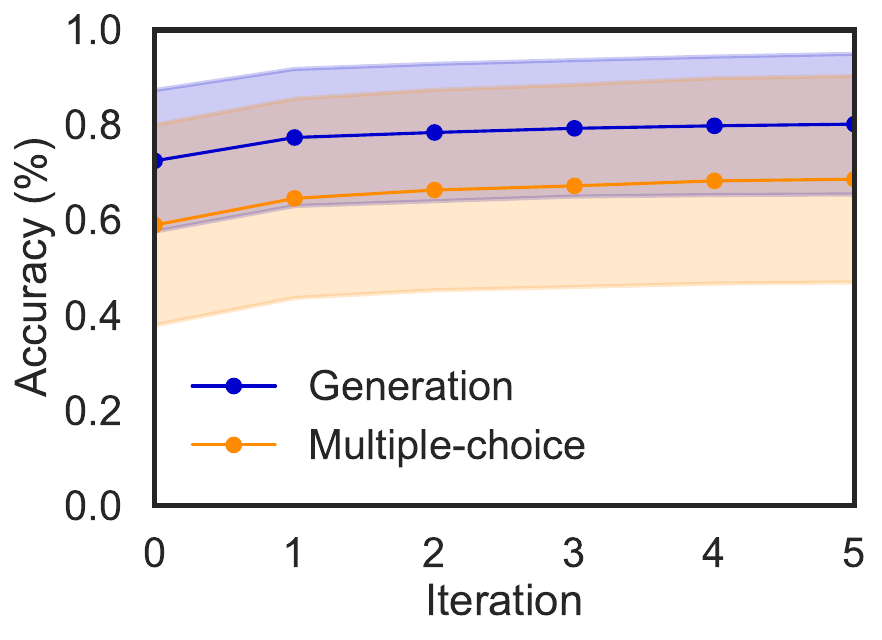}\label{fig:f1}}
  \caption{\textbf{Average cumulative accuracy on generation and multiple-choice}. (\textbf{Top}) Accuracy on the \bbehqa dataset shows that models perform better on the multiple-choice task when we iteratively self-correct the model response to the questions, while (\textbf{bottom}) shows the accuracy on the \tinytruthfulqa dataset, indicating that models perform better in generation tasks.}
  \label{figure:average_cum_accuracy}
\end{figure*}

\paragraph{Research Questions.}
Our study is guided by the following three research questions:

\begin{itemize}
\item \textbf{RQ1}: How do self-correction dynamics differ between open-ended and multiple-choice tasks?
\item \textbf{RQ2}: How do model scale and prompting strategy influence self-correction across the two paradigms?
\item \textbf{RQ3}: How does iterative self-correction affect correctness, stability, and semantic drift, and what mechanisms explain these effects?
\end{itemize}

\paragraph{Datasets.}
\label{sec:dataset}
We evaluate on two benchmarks, \bbehqa and \tinytruthfulqa, that each provide parallel formulations for both multiple-choice questions and open-ended generation. This allows us to study self-correction dynamics under consistent task content but different output constraints.

\begin{itemize}
    \item \bbehqa \cite{kazemi2025big} is typically phrased in multiple-choice form, where each question presents a pronoun or reference with referential ambiguity and provides four candidate referents. However, the same questions can also be cast into an open-ended format by asking models to generate the referent rather than choose among options. Thus, \bbehqa instantiates a scenario where the answer space is tightly constrained but also amenable to open-ended generation in a parallel setup.
    \item \tinytruthfulqa \cite{polo2024tinybenchmarks} is a challenging subset of the TruthfulQA benchmark \cite{lin2022truthfulqa} focused on short-form factual queries that tend to provoke false or misleading answers from LLMs. While TruthfulQA is usually evaluated via free-form generation, where models must produce a truthful answer, a multiple-choice variant has also been developed, offering for each question a small set of candidate answers drawn from the same reference answer pool. Therefore, \tinytruthfulqa inherits this dual-format nature, where the same questions support both open-ended and multiple-choice instantiations. This dataset exemplifies scenarios requiring knowledge retrieval and precision in generation.
\end{itemize}

By evaluating both tasks, we cover one case where the ground-truth answer is within a closed set of options and one case where the answer must be generated. We therefore can compare how iterative self-correction dynamics differ when the model's output is tightly constrained versus freely generative.

\paragraph{Models.}
\label{sec:models}
We evaluate the dynamics of iterative self‐correction under unconstrained generation and multiple‐choice selection using six pre‐trained language models ranging from small to large parameters. We evaluate \smol \cite{allal2025smollm}, \qwenthree \cite{qwen2025qwen25technicalreport}, \llama \cite{grattafiori2024llama}, \qwenfourteen \cite{qwen2025qwen25technicalreport}, \deepseek \cite{guo2025deepseek}, and \gemini \cite{comanici2025gemini}. These models represent diverse families and scales (from distilled smaller models to state-of-the-art large models). For each model and dataset, we compare three aforementioned prompting strategies: a direct \baseline prompt, zero‐shot chain‐of‐thought (\chainot) prompting \cite{kojima2022large}, and our iterative \selfconsistency procedure that reviews and refines the model's own previous response for \textit{up to five rounds}. We use HuggingFace to run the models except \gemini, which is accessed through the API.

\paragraph{Prompts.}
\label{sec:prompts}
In our experiments, we use simplified prompts to minimize the impact of prompt design on performance across tasks, keeping the focus on the self-correction mechanism \cite{huang2023cannot-self-correct}. Specifically, we apply a basic prompt for the \baseline method and adopt zero-shot Chain-of-Thought (\chainot) prompting \cite{kojima2022large} for both the \chainot and Self-Consistency (\selfconsistency) approaches. The initial prompts are used for the first attempt (iteration 0) under each strategy. They differ only in whether the model is encouraged to produce an explicit chain of reasoning before the final answer. For iterations beyond the first, we prepend instructions to review the prior attempts. In both cases, the model is reminded of its earlier answers (which are included in the conversation context) and encouraged to refine them. The \chainot variant additionally maintains the directive to use a step-by-step reasoning process during revision. Our full prompts can be found in Appendix \ref{sec:app:prompts}.

\paragraph{Final Answer Extraction.}
\label{sec:final-answer}
For all of our problems, we added the `\textbf{The final answer is: }' suffix to the text of the prompt to encourage the model to produce the final answer in a format that we can easily extract. More details in Appendix \ref{sec:app:final-answer}.
\section{Results}
\label{sec:results}

\begin{figure}[!tbp]
  \centering
  \subfloat[Baseline]{\includegraphics[width=0.32\columnwidth]{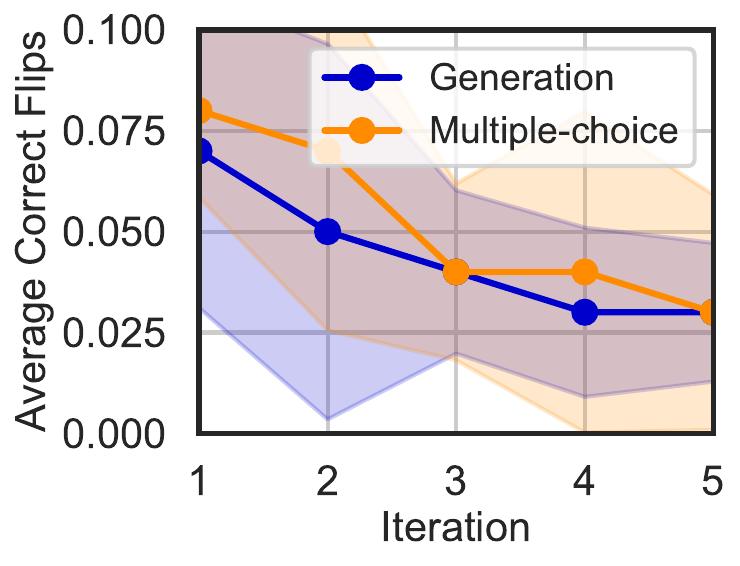}\label{fig:f1}}
  \subfloat[CoT]{\includegraphics[width=0.32\columnwidth]{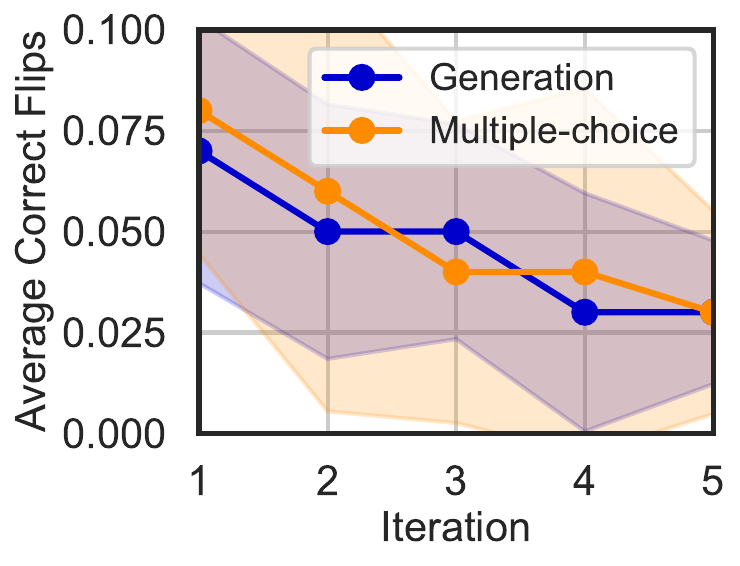}\label{fig:f2}}
  \subfloat[\selfconsistency]{\includegraphics[width=0.32\columnwidth]{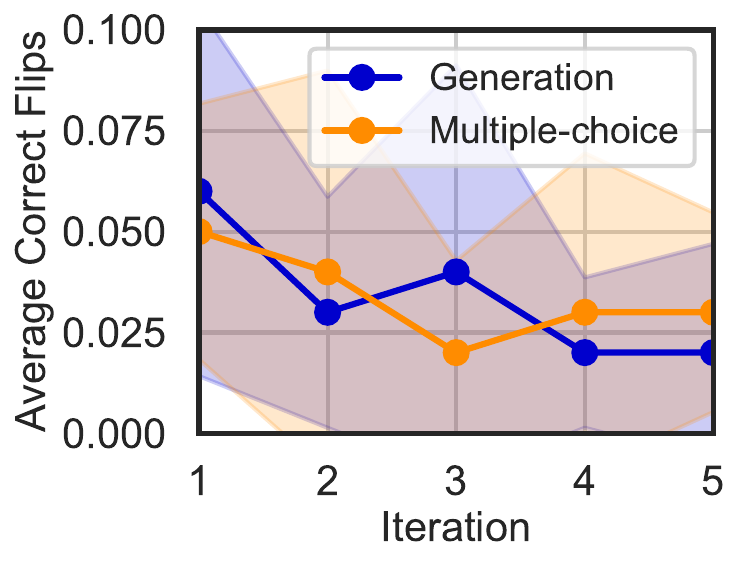}\label{fig:f2}}
  \qquad
    \subfloat[Baseline]{\includegraphics[width=0.32\columnwidth]{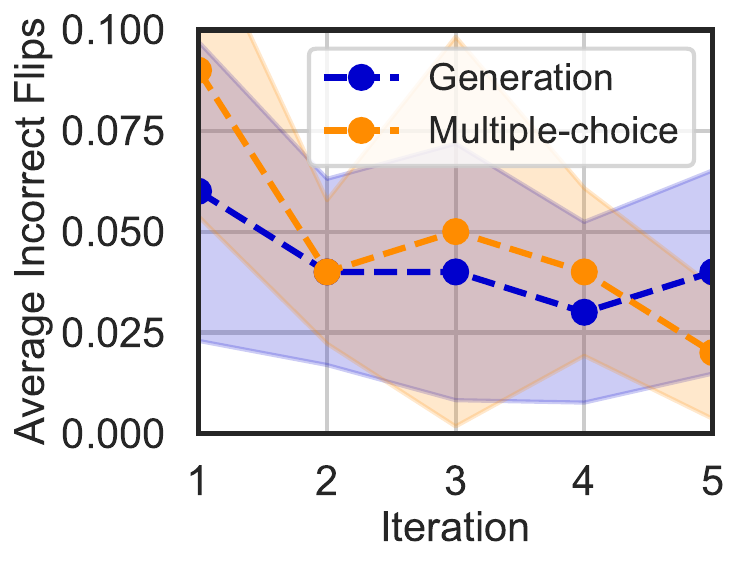}}
  \subfloat[CoT]{\includegraphics[width=0.32\columnwidth]{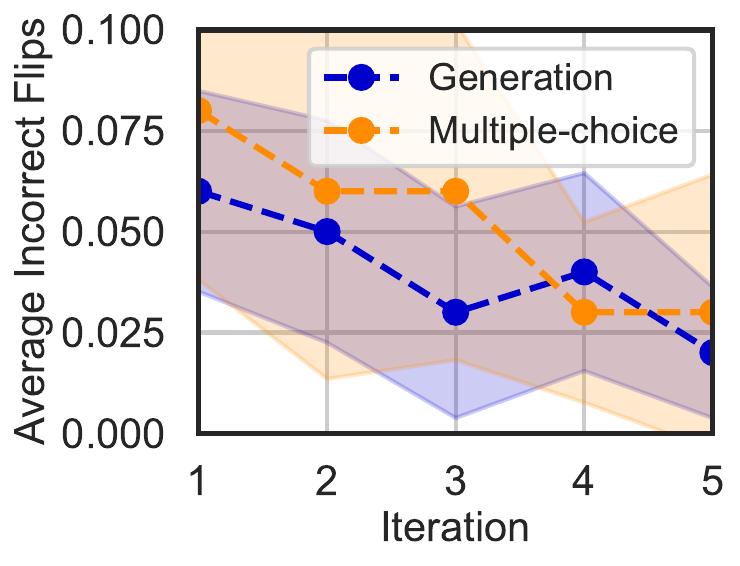}\label{fig:f2}}
  \subfloat[\selfconsistency]{\includegraphics[width=0.32\columnwidth]{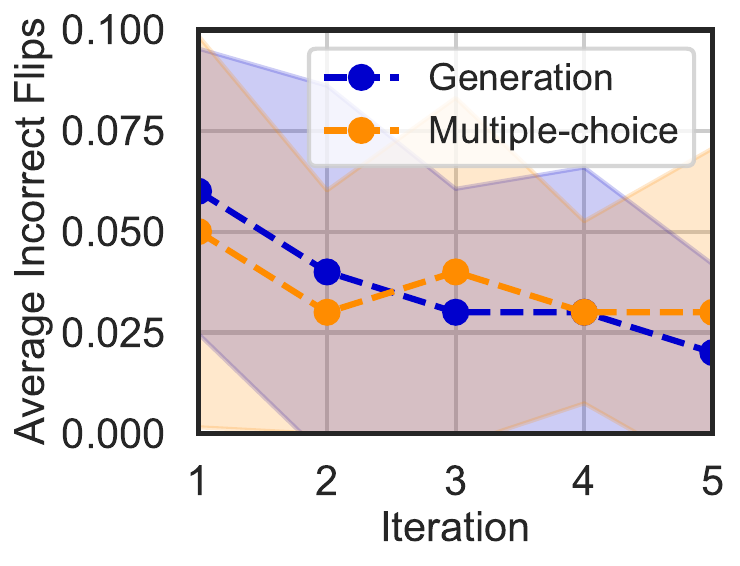}\label{fig:f2}}
  \caption{Average Correct and Incorrect Flips on \bbehqa}
  \label{fig:average_flip_bbehqa}
\end{figure}

We now analyze the results in relation to our three research questions.

\paragraph{Improvement Patterns Across Iterations (RQ1).}
To address RQ1, we first examine the aggregate performance reported in Figure \ref{figure:average_cum_accuracy}, which compares accuracy across correction iterations for generation and multiple-choice formats. The generation paradigm improves rapidly in the first one or two iterations, showing that early revisions are effective at fixing obvious errors or adding missing information. However, after these early gains, performance often plateaus or declines, as additional revisions increase the risk of semantic drift and lead to new mistakes. In contrast, the multiple-choice paradigm improves more gradually and steadily. Accuracy rises incrementally with each round of self-correction, reflecting cautious re-weighting among fixed options. Yet this format struggles to recover from poor initial predictions: if the model’s first choice is wrong, subsequent iterations rarely flip it to the correct option, showing the effects of \textit{logit inertia}.

\begin{figure}[!tbp]
  \centering
  \subfloat[Baseline]{\includegraphics[width=0.32\columnwidth]{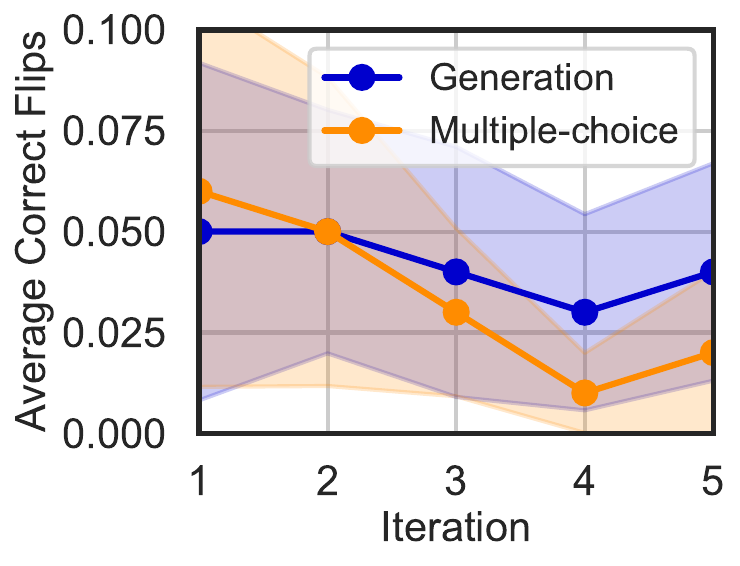}\label{fig:f1}}
  \subfloat[CoT]{\includegraphics[width=0.32\columnwidth]{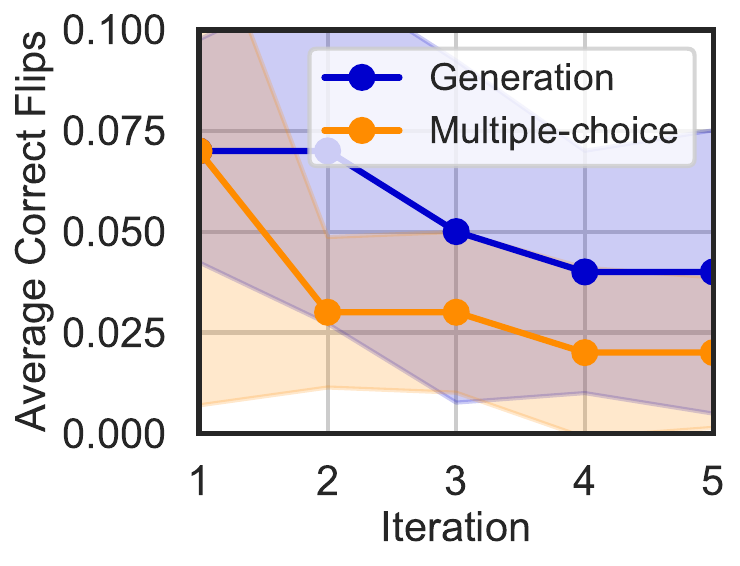}\label{fig:f2}}
  \subfloat[\selfconsistency]{\includegraphics[width=0.32\columnwidth]{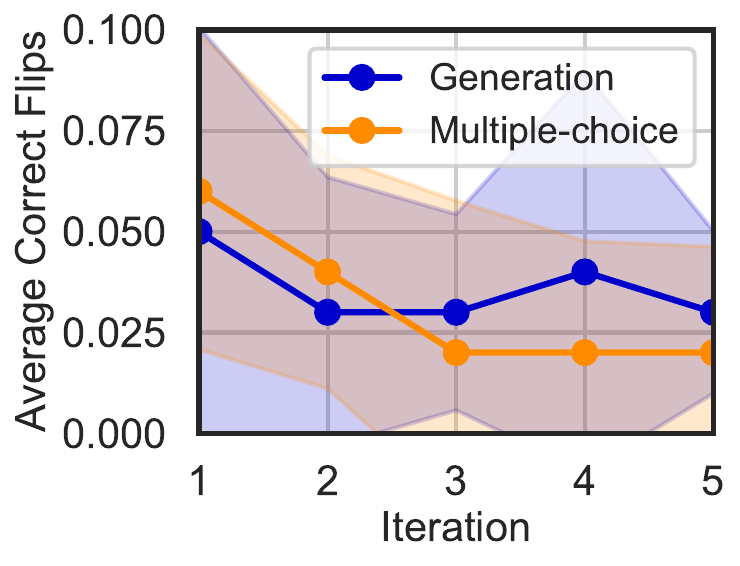}\label{fig:f2}}
  \qquad
    \subfloat[Baseline]{\includegraphics[width=0.32\columnwidth]{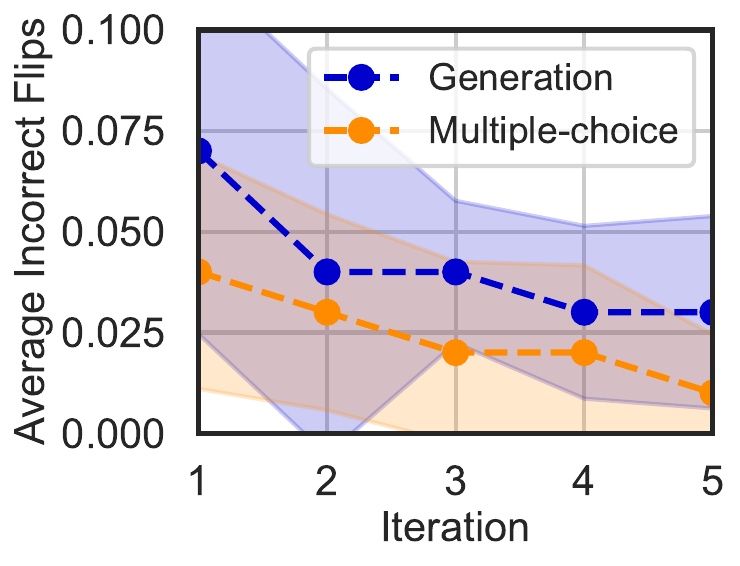}}
  \subfloat[CoT]{\includegraphics[width=0.32\columnwidth]{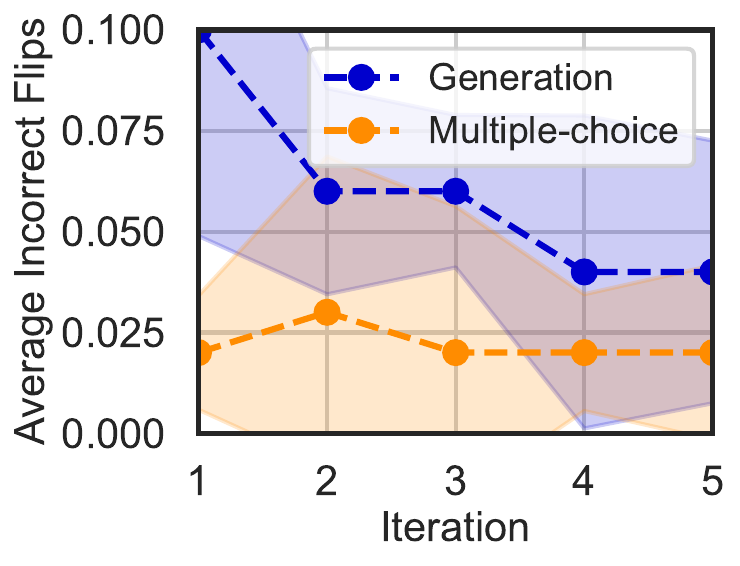}\label{fig:f2}}
  \subfloat[\selfconsistency]{\includegraphics[width=0.32\columnwidth]{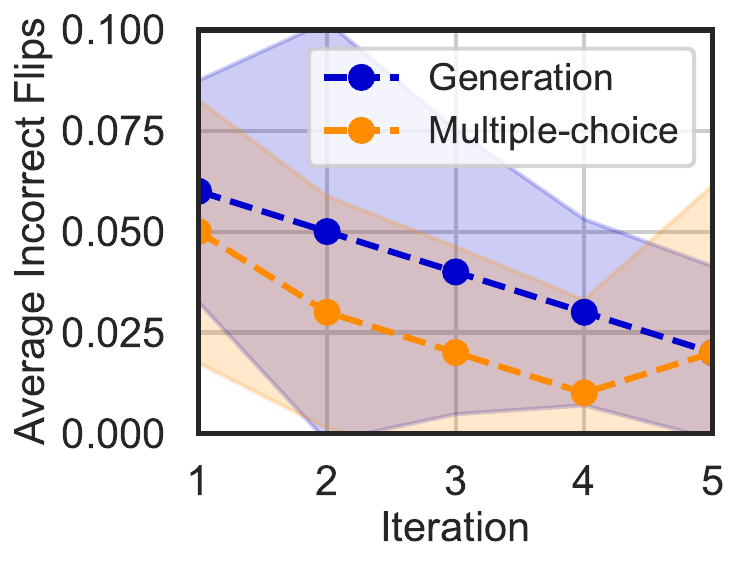}\label{fig:f2}}
  \caption{Average Correct and Incorrect Flips on \tinytruthfulqa}
  \label{fig:avg_flip_tinyqa}
\end{figure}

Figures \ref{fig:average_flip_bbehqa} and \ref{fig:avg_flip_tinyqa} present the ``flip'' dynamics of self-correction on the two datasets, broken down into correct (a previously wrong answer corrected to right) and incorrect (a previously correct answer changed to wrong) flips over successive iterations. On \bbehqa (Figure \ref{fig:average_flip_bbehqa}), multiple-choice self-correction yields very few flips overall. Correct answers are stably retained, but wrong initial guesses are seldom corrected. Generation, by contrast, produces more frequent flips: many beneficial in early iterations (correcting ambiguous references) but increasingly harmful in later ones, as correct answers are sometimes replaced with incorrect ones, once the model starts to over-correct or drift. On \tinytruthfulqa (Figure \ref{fig:avg_flip_tinyqa}), the contrast is sharper: generation produces a high number of flips, with many early correct flips (replacing misconceptions with truths), but also a rising number of incorrect flips in later rounds, reflecting semantic drift. Multiple-choice again remains stable, with minimal incorrect flips but limited ability to recover from an early mistake. 

Taken together, we show that open-ended generation offers adaptability and rapid early gains but suffers from instability in later iterations, whereas multiple-choice offers stability and incremental improvement but is hampered by inertia when the first choice is wrong. This confirms that self-correction effectiveness is strongly dependent on task format: open-ended generation can exploit flexibility to correct errors but risks drift, while multiple-choice provides reliable retention of correct answers at the expense of recoverability. If the model doesn’t get the answer right on the first attempt, it has a hard time changing to the correct option later. This fundamental difference in dynamics directly answers RQ1: self-correction behaves very differently in open-ended versus fixed-option scenarios, with each paradigm exhibiting its own pattern of improvement and failure modes.

\paragraph{Effects of Model Scale and Prompting Strategy (RQ2).}

\begin{figure*}
    \centering
    \includegraphics[scale=0.31]{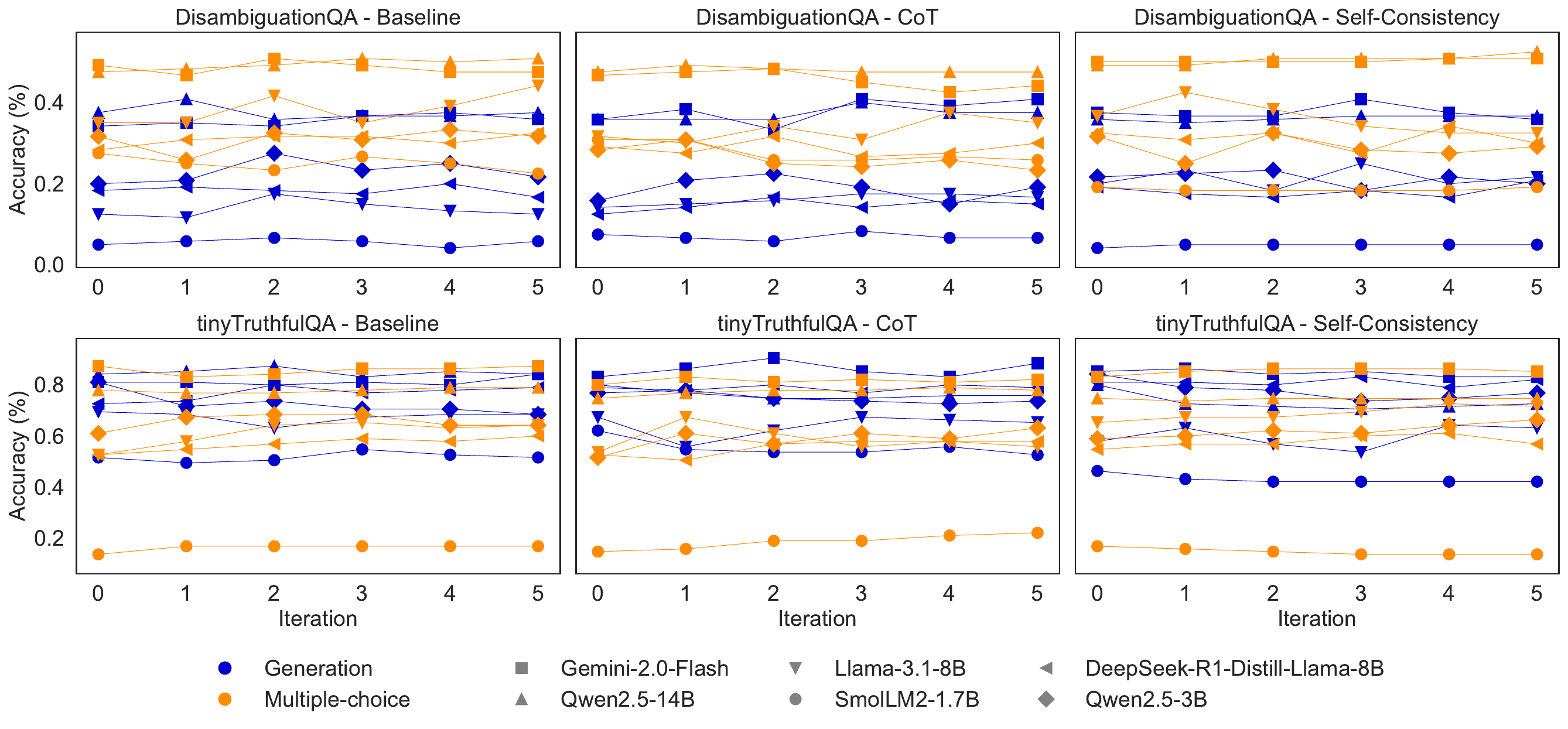}
    \caption{\textbf{Accuracy per iteration per model on generation and multiple-choice.}}
    \label{fig:all-acc-iter}
\end{figure*}

Here, we investigate how a model’s size and the prompting strategy influence self-correction, and whether these effects differ between the two output paradigms. Figure \ref{fig:all-acc-iter} provides a detailed view of accuracy per iteration for various models under different prompting methods. A clear finding is that task difficulty moderates these effects. On the challenging \bbehqa benchmark, accuracy is low for all models: even the largest (e.g., \gemini, Qwen2.5-14B) plateau around 50\% in multiple-choice and below 20\% in generation, while smaller models perform far worse. In contrast, on the easier \tinytruthfulqa, generative accuracy ranges from 60–90\% and multiple-choice from 50–80\%, with even small models performing well. Thus, model scale yields clear benefits on harder tasks, but differences narrow considerably on simpler ones.

The prompting strategy has a modest but noticeable effect, more so on the difficult task. On \bbehqa, using an explicit \chainot prompt or a \selfconsistency approach yields slight accuracy improvements over the \baseline direct prompting. For example, prompting the model to “think step by step” or to consider multiple reasoning paths sometimes helps it disambiguate the question better, nudging up the accuracy by a few percentage points. These gains, while not dramatic, suggest that reasoning-oriented prompts can aid the model on ambiguous, challenging questions. In contrast, on \tinytruthfulqa, all three prompting strategies lead to very similar performance. The accuracy curves for different prompts on this task are nearly overlapping (Figure \ref{fig:all-acc-iter}), indicating that when a question is relatively straightforward or the model already knows the domain (e.g., common truths vs. misconceptions), an elaborate prompt does not provide much benefit. In summary, prompting variations have a task-dependent impact: they can be slightly beneficial for resolving difficult queries (\bbehqa) but mostly redundant for simpler factual questions (\tinytruthfulqa). This aligns with the findings in the literature \cite{spraguecot}.

Model scale shows a similarly nuanced effect. Larger models generally outperform smaller ones, especially on \bbehqa, where 14B+ models clearly surpass 1–3B models. On \tinytruthfulqa, however, the performance gap narrows, with small models often approaching large-model accuracy. In some cases, scaling produces diminishing returns, indicating that size matters more for difficult tasks but offers limited advantage once a task is already within reach.

Notably, repeated iterations of self-correction do not consistently boost accuracy for either paradigm, regardless of model size or prompt strategy. Across our experiments, most performance curves over iterations (spanning iteration 0 through 5) are relatively flat after the initial step. As highlighted by Figure \ref{fig:all-acc-iter}, it is rare to see a clear upward trajectory beyond the first one or two iterations; instead, accuracy often oscillates with minor gains or losses. For example, a model might correct a mistake at iteration 1, only to introduce a different mistake at iteration 3, ending up with a similar accuracy as it started. This plateauing behavior implies that giving the model many chances to self-correct yields diminishing returns. Neither larger scale nor advanced prompting fundamentally changes this outcome – their benefits tend to manifest in the first attempt or two, but they do not drive continual improvement with more iterations. In some cases, we even observed slight performance degradation with too many iterations (echoing the drift issues from RQ1). In summary, the impact of model scale and prompting strategy on self-correction is real but nuanced: larger models and CoT-style prompts can improve initial accuracy, especially on hard tasks, but these factors are task-dependent and ultimately insufficient to guarantee ongoing improvements through iterative self-correction alone. Multiple-choice and generation formats alike see their gains saturate early, and improvements from scaling or better prompting taper off without addressing the core limitations of each paradigm. Notably, we also found that the multiple-choice paradigm often reaped slightly more benefit from increased model size and reasoning prompts than the generation paradigm did (especially on \bbehqa), reinforcing the idea that constrained decision tasks can more readily capitalize on those enhancements. Still, neither paradigm achieves a dramatically upward performance trend with iteration – a key insight for understanding the boundaries of current self-correction capabilities.

\paragraph{Trade-offs Between Adaptability and Stability (RQ3).}
RQ3 examines how iterative self-correction influences correctness, stability, and semantic drift across unconstrained and constrained outputs. In the generation setting, flexibility allows models to revise and often improve answers in the first one or two iterations, but this same flexibility leads to semantic drift in later rounds. As Figures \ref{fig:average_flip_bbehqa} and \ref{fig:avg_flip_tinyqa} as well as the detailed plots of per model evaluation in Appendix \ref{sec:app:flips}, generation produces many flips: early ones are often correct (e.g., resolving an ambiguity or correcting a misconception), but over time, incorrect flips dominate as the model over-edits or drifts away from the question. This suggests that while generation supports adaptability, it lacks effective internal checks to prevent harmful revisions. By contrast, in the multiple-choice setting, the output space is restricted to fixed options, which prevents drift altogether. Correct answers remain locked in across iterations, reflecting high stability. However, this comes with logit inertia: wrong initial answers persist, with very few corrective flips observed in Figures \ref{fig:average_flip_bbehqa} and \ref{fig:avg_flip_tinyqa}. The mechanism here is that once a wrong option is selected, the model rarely shifts its ranking enough to choose the correct one later, even when revisiting its reasoning.

These patterns reveal a fundamental adaptability–stability trade-off. Generation is exploratory and can recover from initial mistakes, but risks undermining correctness as iterations accumulate. Multiple-choice ensures consistency once correct, but limits opportunities to fix errors. For system design, this implies that neither paradigm is universally optimal. Applications requiring stable outputs, such as safety-critical domains, benefit from constrained correction, though additional mechanisms may be needed to overcome inertia (e.g., external verification or re-ranking). Conversely, tasks where capturing every possible correction is crucial may favor open-ended revision, provided that safeguards against drift are implemented. Promising directions include hybrid strategies that combine paradigms, using generation to explore candidate answers followed by constrained verification to anchor correctness, and dynamic stopping rules that halt iteration once improvements saturate or harmful drift is detected. Addressing these trade-offs directly, by mitigating semantic drift in generation and reducing inertia in multiple-choice, will be key to making iterative self-correction a reliable capability of LLM systems.
\section{Conclusion}
\label{sec:conclusion}
This study compared iterative self-correction in large language models across open-ended generation and multiple-choice question answering. Results show that the structure of the output space fundamentally shapes correction dynamics. Generation achieves rapid early gains by correcting errors in the first few iterations, but suffers from semantic drift as revisions accumulate, resulting in increasing rates of incorrect flips. Multiple-choice responses remain highly stable and avoid drift, but exhibit \textit{logit inertia}: wrong initial answers are rarely overturned, and improvements are incremental at best. Model scale and prompting strategy modulate performance but do not alter these core patterns. Larger models and reasoning-oriented prompts (\chainot, \selfconsistency) yield slight improvements, especially on the harder \bbehqa task, but their effects are modest and task-dependent. Across both paradigms, accuracy generally plateaus after the first one or two iterations, showing that repeated self-correction brings limited benefit.

These findings highlight an inherent adaptability–stability trade-off. Open-ended generation enables recovery from errors but risks instability, while multiple-choice ensures reliability but limits correction. Future work should explore hybrid strategies, such as using generation for exploration and constrained formats for verification, as well as dynamic stopping criteria to prevent late drift. Addressing drift and inertia directly will be essential for building reliable self-correcting LLM systems.

\newpage

\section*{Limitations}
\label{sec:limit}
This study focuses on benchmarks that provide parallel formulations for both open-ended generation and multiple-choice questions. While this setup enables a controlled analysis of self-correction across task formats, it also limits the number of datasets available for evaluation, as few benchmarks support both types of tasks. Moreover, our experiments are conducted using currently available models of moderate scale. Recent larger models, which may exhibit different self-correction dynamics and reasoning behaviors, are not included in our analysis. Future work could extend our study to such models to provide a more comprehensive understanding of scaling effects.

\section*{Ethical Considerations}
We have carefully verified that the software, model checkpoints and existing datasets utilised in this work are permitted for access, distribution and, where relevant, modification. Our use and purpose comply with those terms.

\section*{Acknowledgments}
This research is supported by the Engineering and Physical Sciences Research Council [EP/S021566/1] and the EPSRC Fellowship titled ``Task Based Information Retrieval'' [EP/P024289/1].

\bibliography{references}

\clearpage
\appendix
\section{Details on Experimental Setup}

\subsection{Details on Final Answer Extraction}
\label{sec:app:final-answer}
For all of our problems, we added a short phrase to the text of the question to guide the model to give the final answer in a clear format:
``\textit{provide your final answer after the `The final answer is: '}.'' To extract the answer, we split the output of the model using this phrase and take what comes after it. Since models sometimes change the phrase slightly, we also check for different variations until one is found: ``The answer is: '', ``The answer is ''. ``The final answer is: '', ``The final answer is ''. Once we get the final answer, we clean it up with a few simple steps:
\begin{enumerate}
    \item If the answer is inside symbols like \textit{boxed}, \textit{text}, \texttt{texttt}, or wrapped in **, we remove those and keep only the text inside.
    \item For multiple-choice questions, if the model adds extra text after the final answer (for example, by putting a newline \verb|\n|), we split on \verb|\n| and keep only the first part. We then lowercase both the final answer and the label, and then check the correctness with the following rules:
    \begin{itemize}
        \item If the final answer and label are identical, we consider the final answer correct.
        \item If they only differ by quotes or brackets around the answer, we consider it to be correct.
        \item For multiple-choice questions, the label is in the format (<LETTER>). If the model only gives the letter (like A instead of (A)), we still count it as correct.
    \end{itemize}
\end{enumerate}

\subsection{Prompts}
\label{sec:app:prompts}

\subsubsection{Start Prompts}

\begin{custombox}{Baseline}
Question: \{question\}. Provide your final answer after the `The final answer is: '.
\end{custombox}

\begin{custombox}{Chain-of-Thought (CoT)}
Question: \{question\}. Think step by step, and provide your final answer after the `The final answer is: '.
\end{custombox}

\subsubsection{Iterative (Self-Correction) Prompts}

\begin{custombox}{Baseline}
Question: \{question\}. Review your previous responses, and provide your final answer after the `The final answer is: '.
\end{custombox}

\begin{custombox}{Chain-of-Thought (CoT)}
Question: \{question\}. Review your previous responses, think step by step and provide your final answer after the `The final answer is: '.
\end{custombox}
\section{Evaluation Protocol}
\label{sec:app:evaluation}
Given the differences between task formats, we adopt distinct evaluation strategies tailored to the characteristics of each setting—open-ended generation and multiple-choice questions. For multiple-choice questions, we use Soft Match (SM) \cite{suzgun2024meta, suzgun2025dynamic}, a lenient metric that considers an answer correct if the ground-truth label appears in the model’s output, disregarding minor formatting variations such as punctuation or whitespace.

For open-ended generation, we employ the LLM-as-a-Judge \cite{zheng2023judging} approach to assess the correctness of the generated answers relative to the ground-truth responses for each dataset. Specifically, we use the fine-tuned model\footnote{\url{https://github.com/yizhongw/truthfulqa_reeval}} introduced by \citeauthor{allanai2025judge} for evaluating generations on \tinytruthfulqa. For \bbehqa, we prompt a large model, \textsc{GPT-4o}, by providing the question, the model-generated answer, and the reference answer, asking it to determine whether the generated answer is correct. The exact prompt used for \bbehqa evaluation is shown below:

\begin{custombox}{Evaluation Prompt}
You are an expert in answer correctness evaluation. Given a question, its reference answer, and a generated answer, please evaluate the correctness of the generated answer based on the question and the reference answer. \\

Here are the question, reference answer, and generated answer: \\

- Question: \{\textit{question}\} \\
- Reference Answer: \{\textit{gold answer}\} \\
- Generated Answer: \{\textit{generated answer}\} \\

Please assess the correctness of the generated answer by considering the question and comparing it against the reference answer. \\

Return yes if the generated answer is completely correct, otherwise, return `no'. The final answer must only be `yes' or `no', corresponding to the correctness of the generated answer.
\end{custombox}
\section{Additional Experiments and Results}

\begin{figure*}[!htbp]
  \centering
  \subfloat[Baseline]{\includegraphics[width=0.34\textwidth]{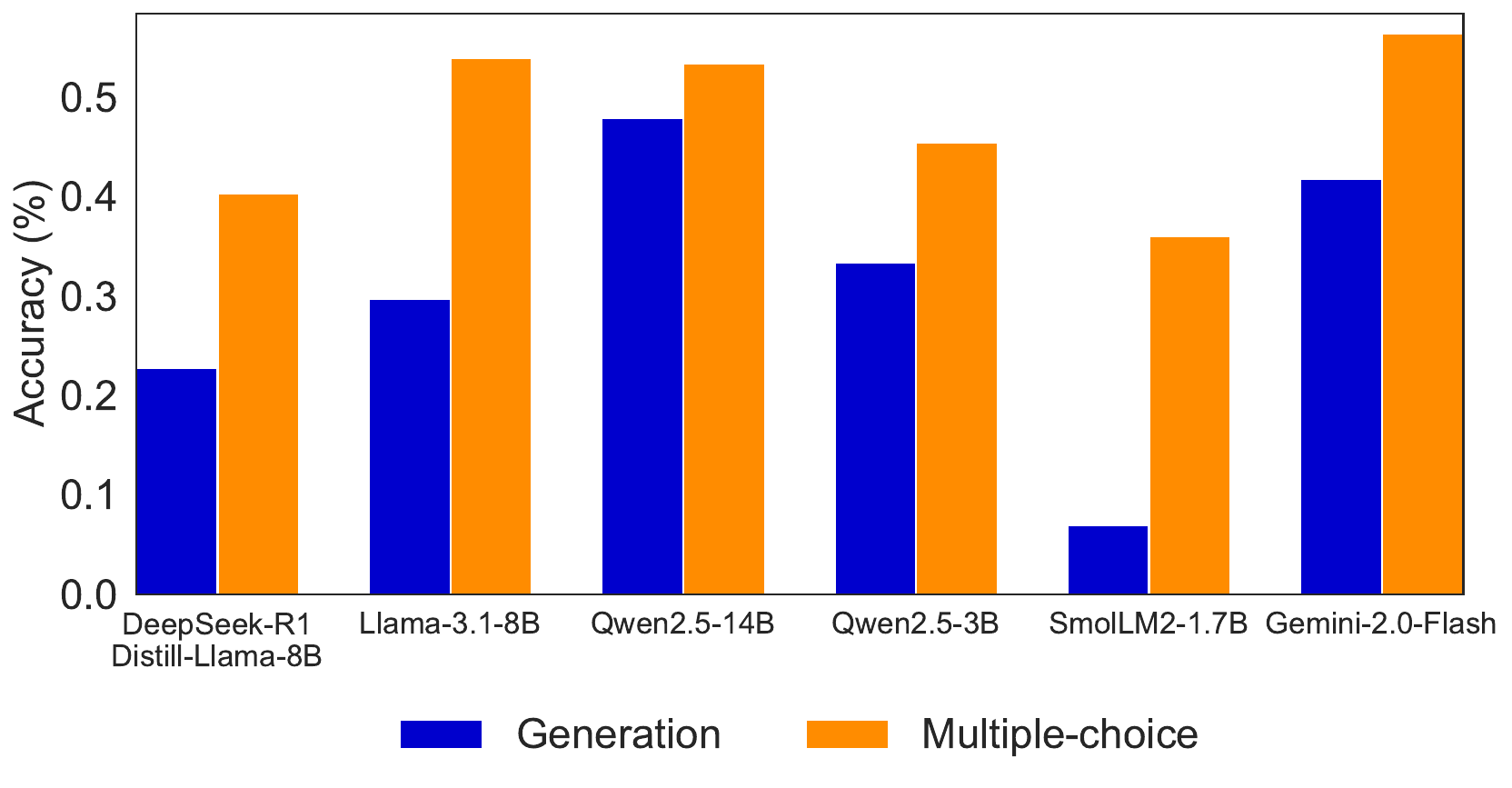}\label{fig:f1}}
  \subfloat[CoT]{\includegraphics[width=0.34\textwidth]{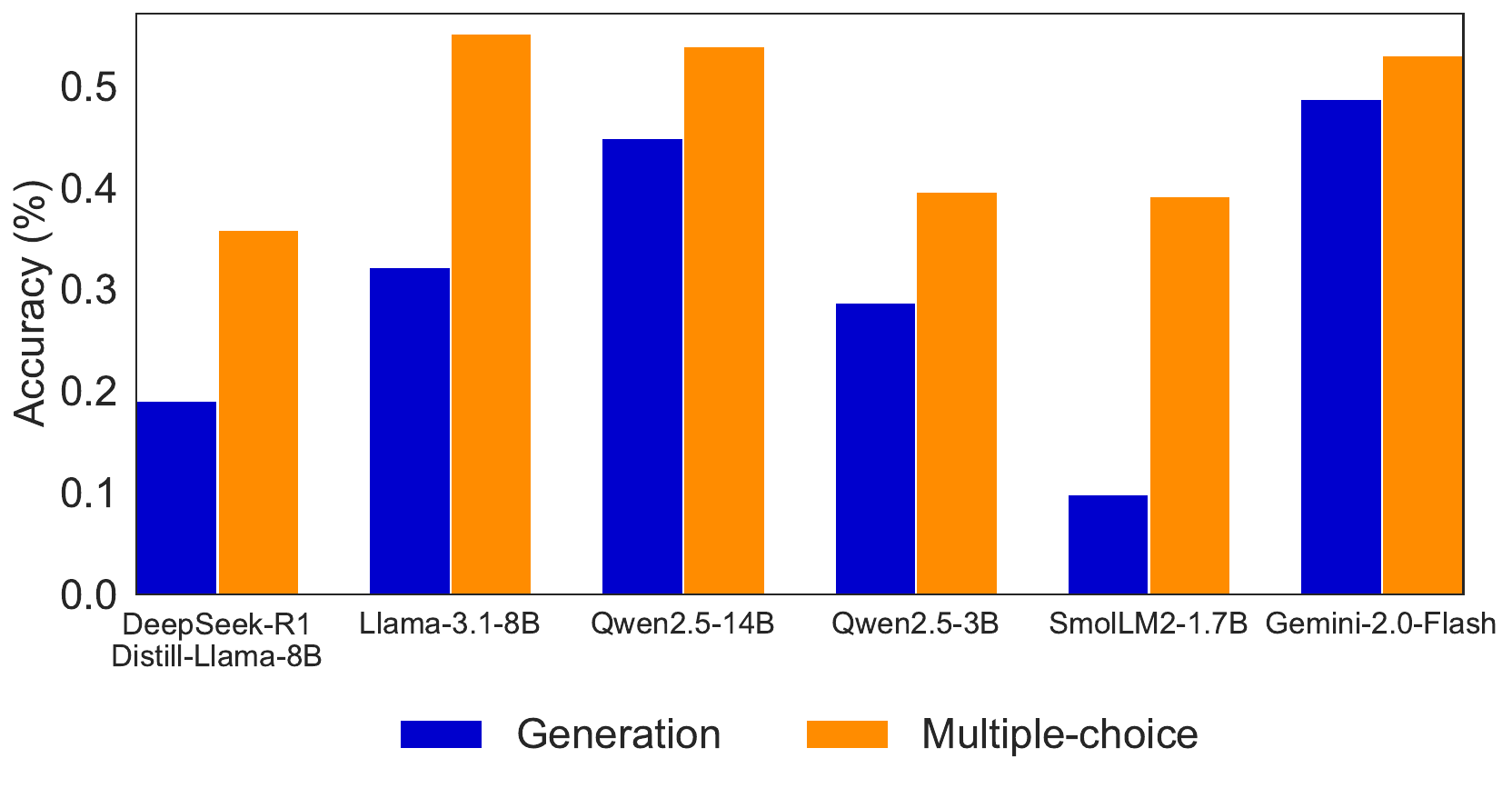}\label{fig:f2}}
  \subfloat[\selfconsistency]{\includegraphics[width=0.34\textwidth]{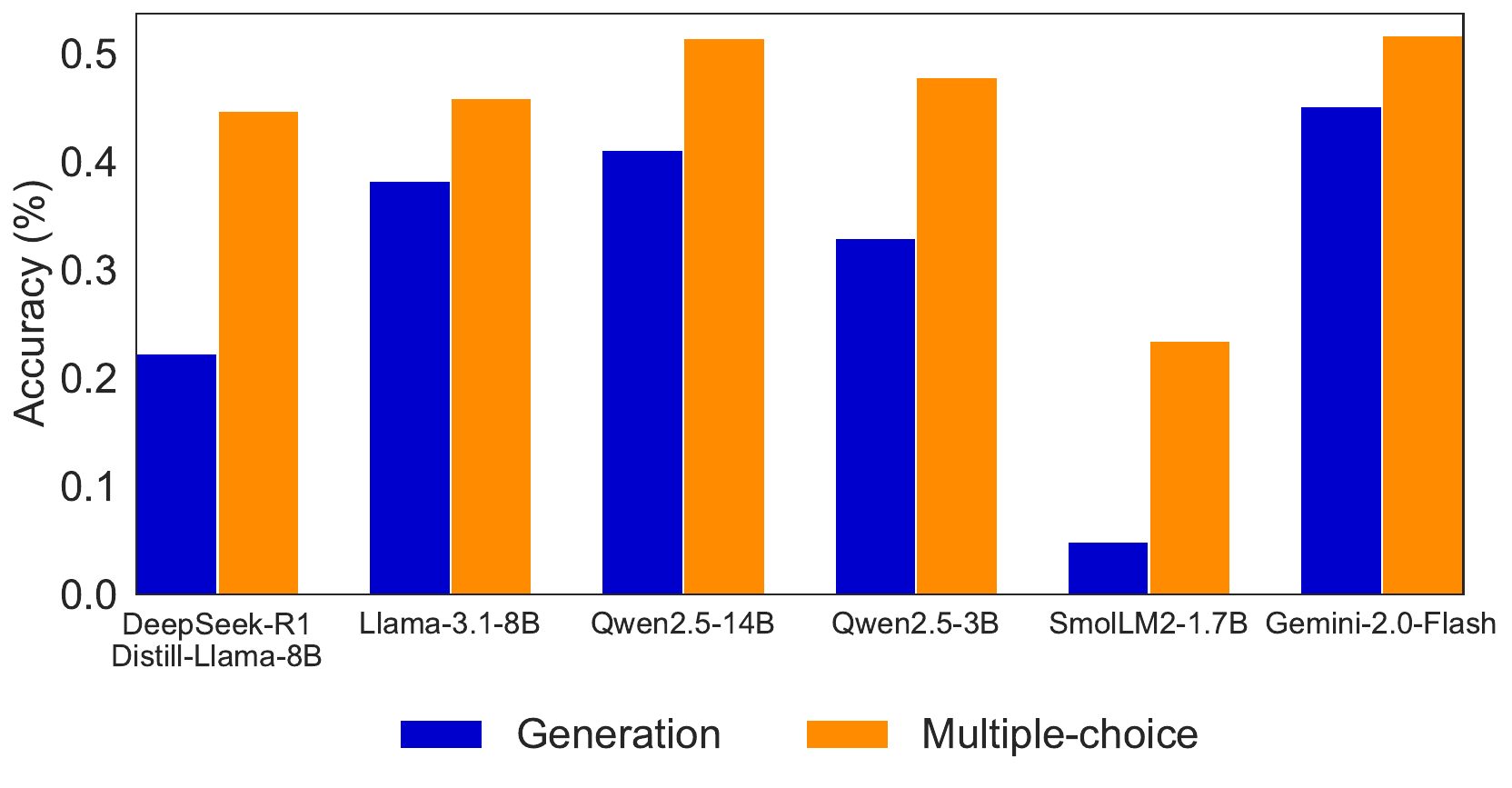}\label{fig:f2}}
  \qquad
   \subfloat[Baseline]{\includegraphics[width=0.34\textwidth]{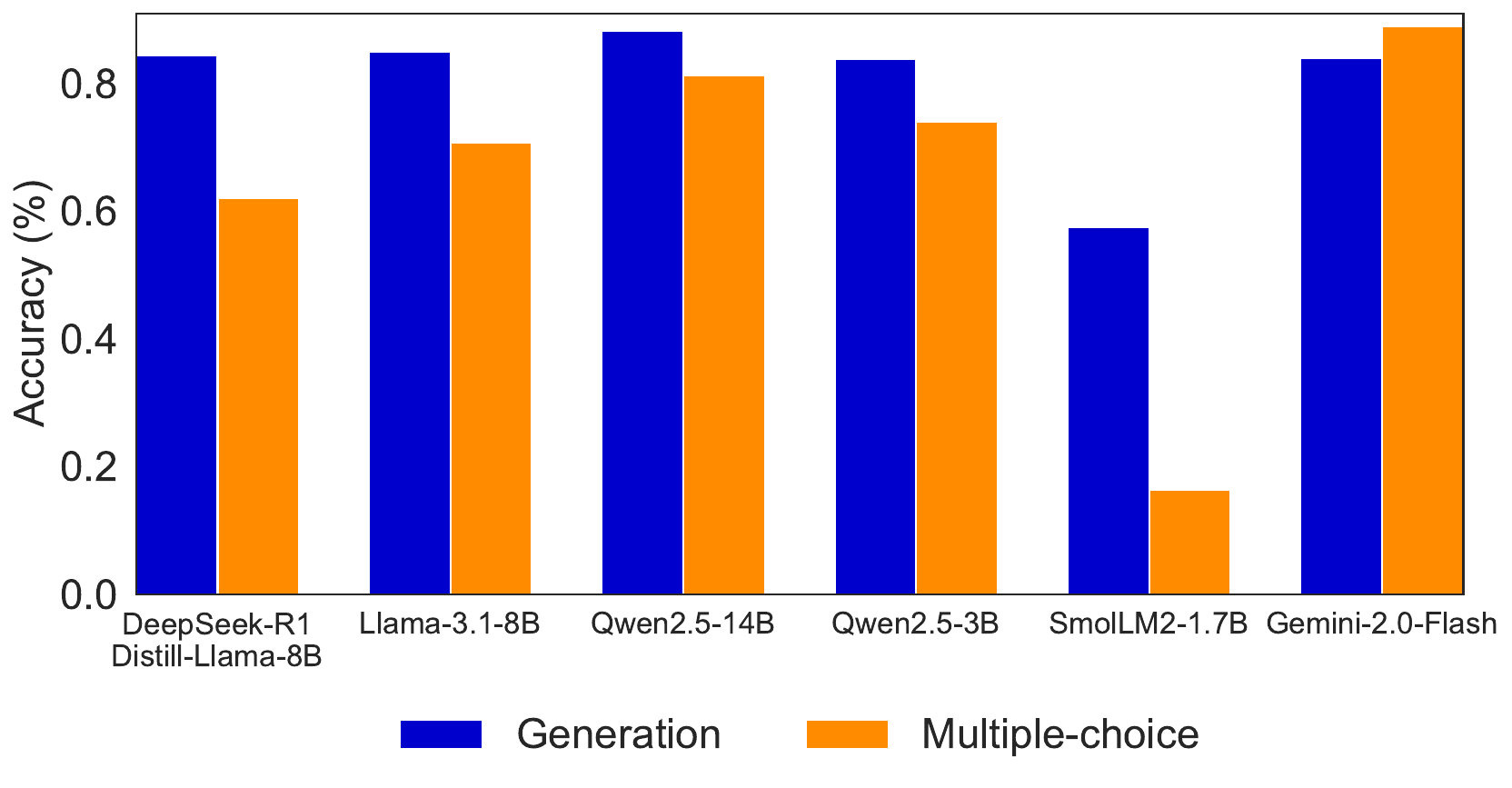}}
  \subfloat[CoT]{\includegraphics[width=0.34\textwidth]{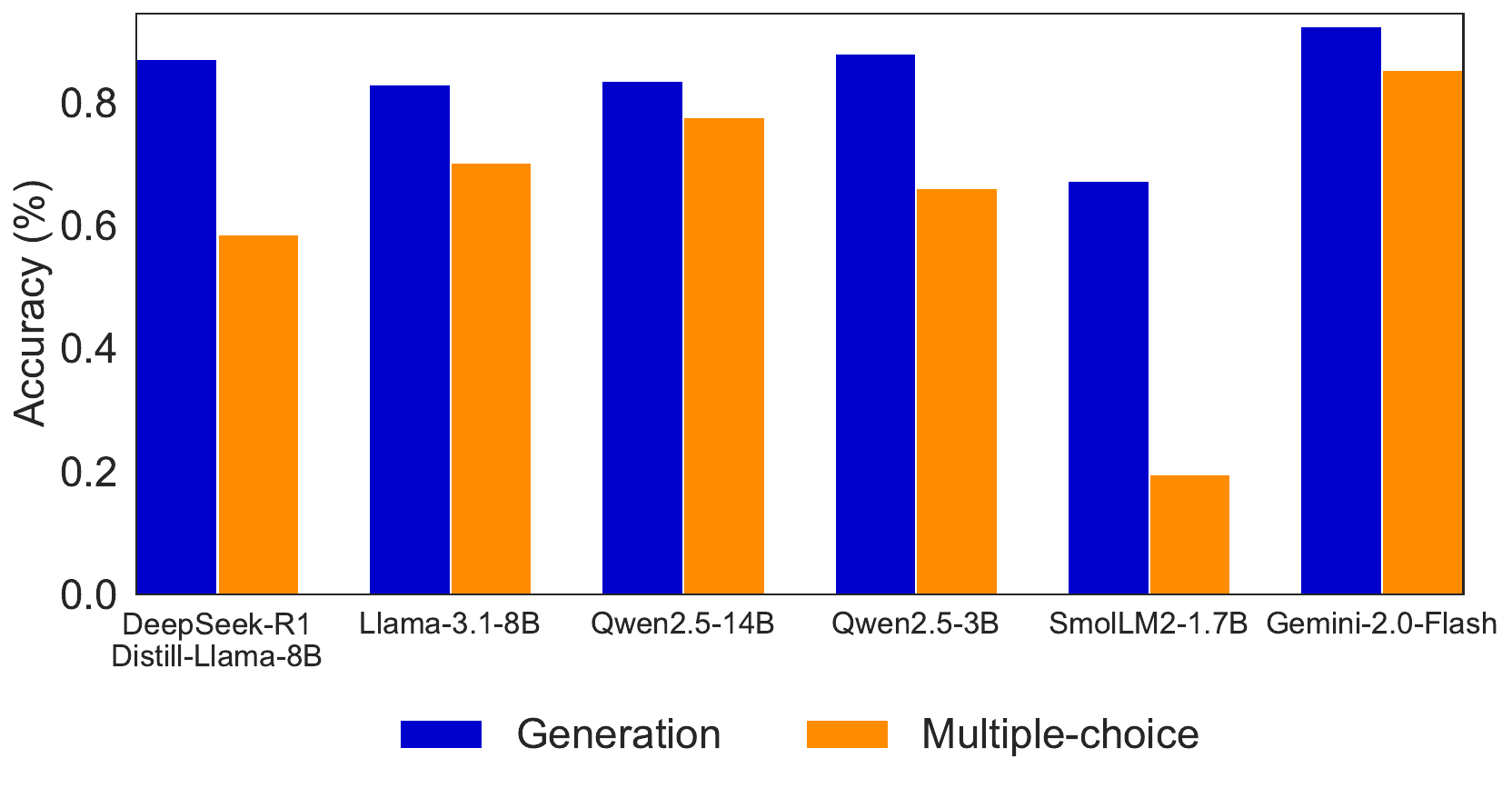}\label{fig:f2}}
  \subfloat[\selfconsistency]{\includegraphics[width=0.34\textwidth]{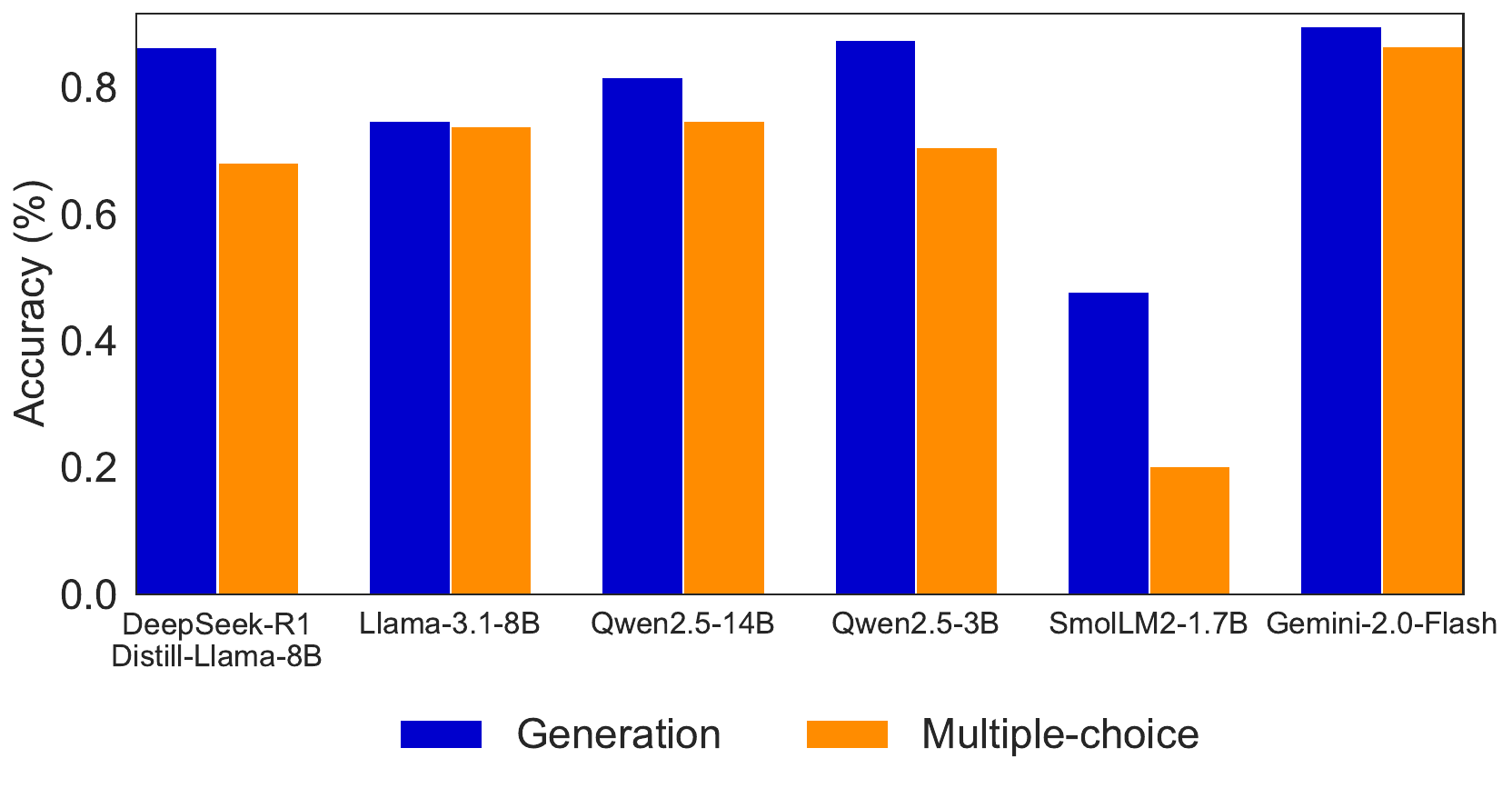}\label{fig:f2}}
  \caption{\textbf{Cumulative accuracy (after final self-correction iteration) using different models on (top) \bbehqa and (bottom) \tinytruthfulqa.} The results indicate that models perform completely differently on self-correction of generation and multiple-choice questions, depending on the dataset.}
  \label{fig:cumulative_accuracy}
\end{figure*}

\subsection{Results on Correct and Incorrect Flips}
\label{sec:app:flips}
Figures 6-11 show the correct and incorrect flips on different datasets and models.


\begin{figure*}[!htbp]
  \centering
  \subfloat[\smol]{\includegraphics[width=0.34\textwidth]{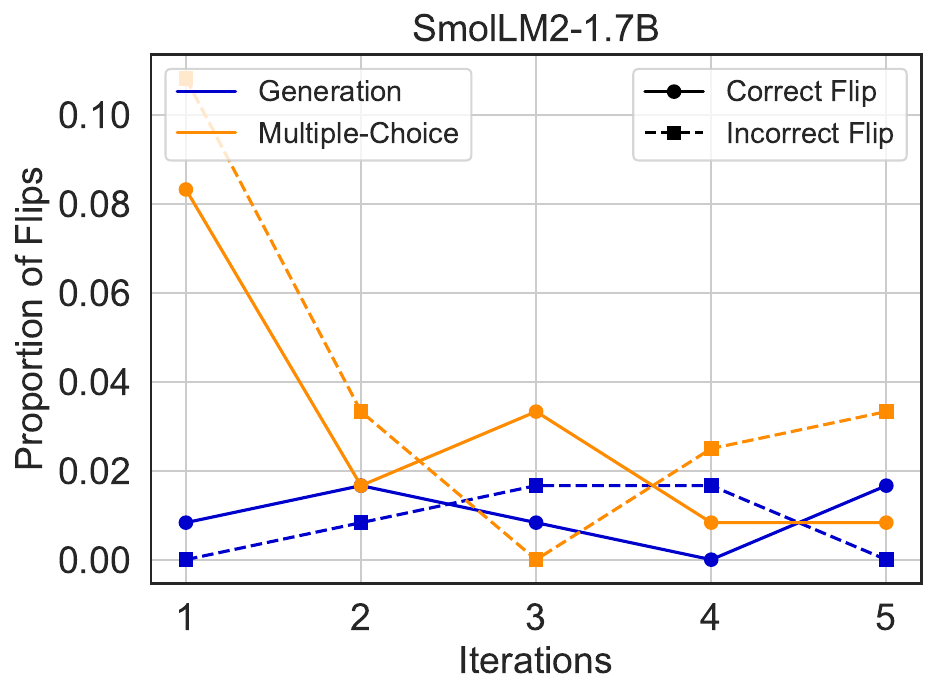}}
  \subfloat[Qwen2.5-3B]{\includegraphics[width=0.34\textwidth]{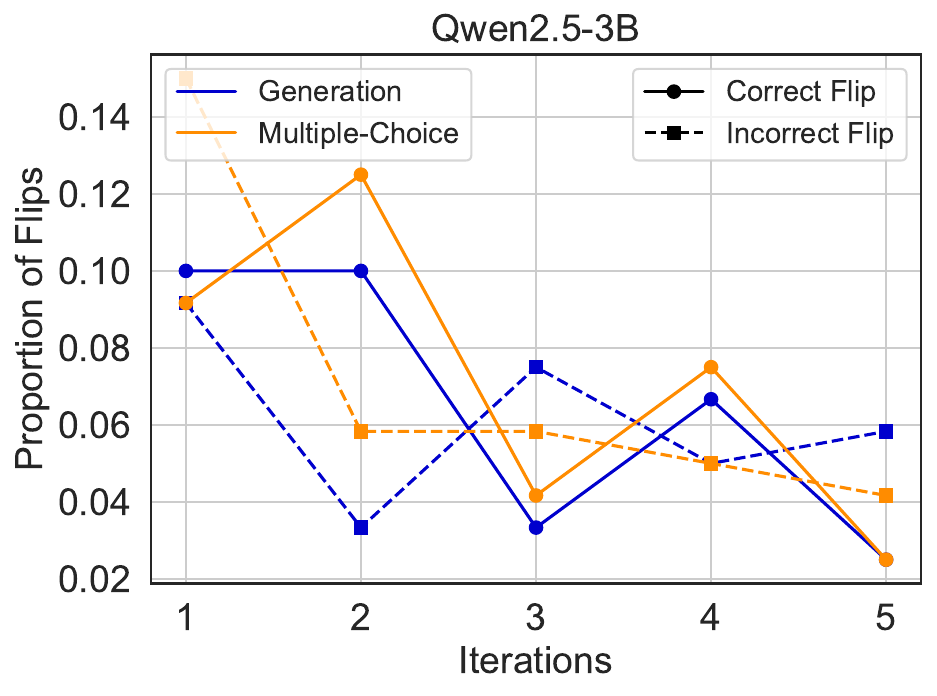}\label{fig:f2}}
  \subfloat[Llama-3.1-8B]{\includegraphics[width=0.34\textwidth]{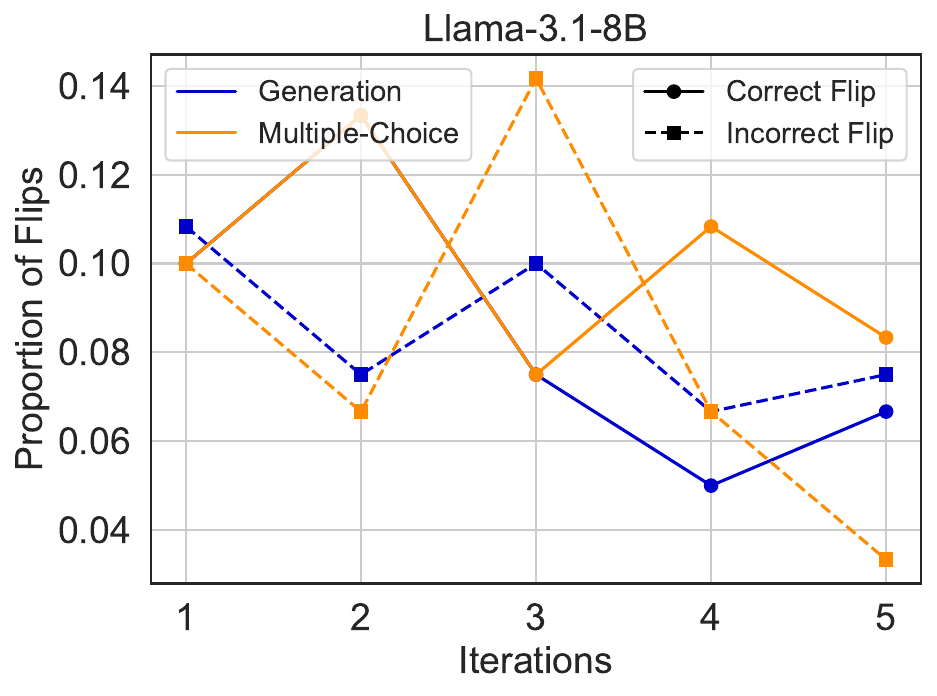}}
  \qquad
    \subfloat[Qwen2.5-14B]{\includegraphics[width=0.34\textwidth]{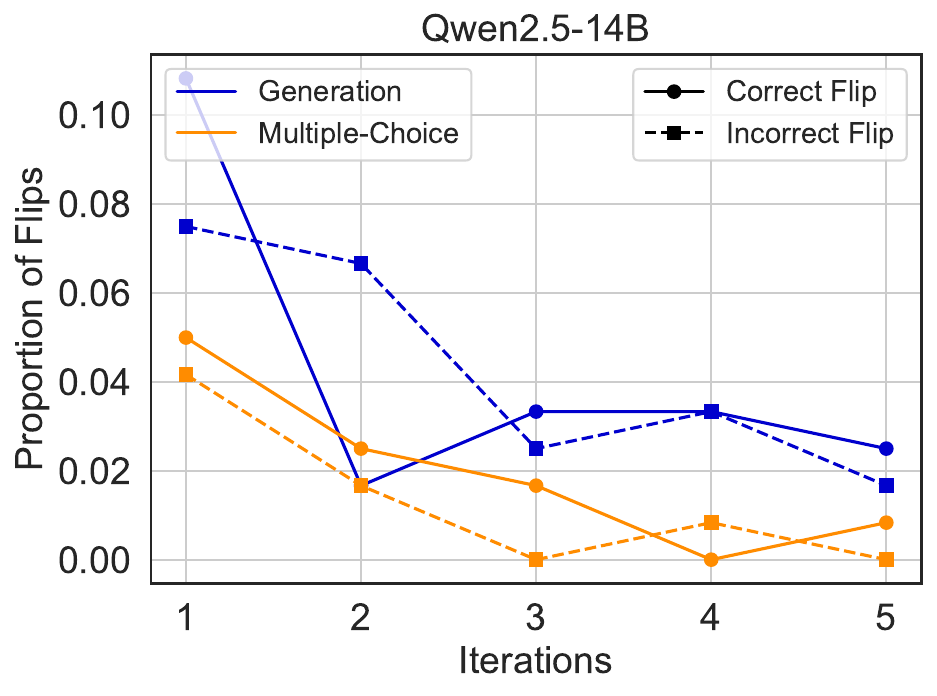}}
  \subfloat[DeepSeek-R1-Distill-Llama-8B]{\includegraphics[width=0.34\textwidth]{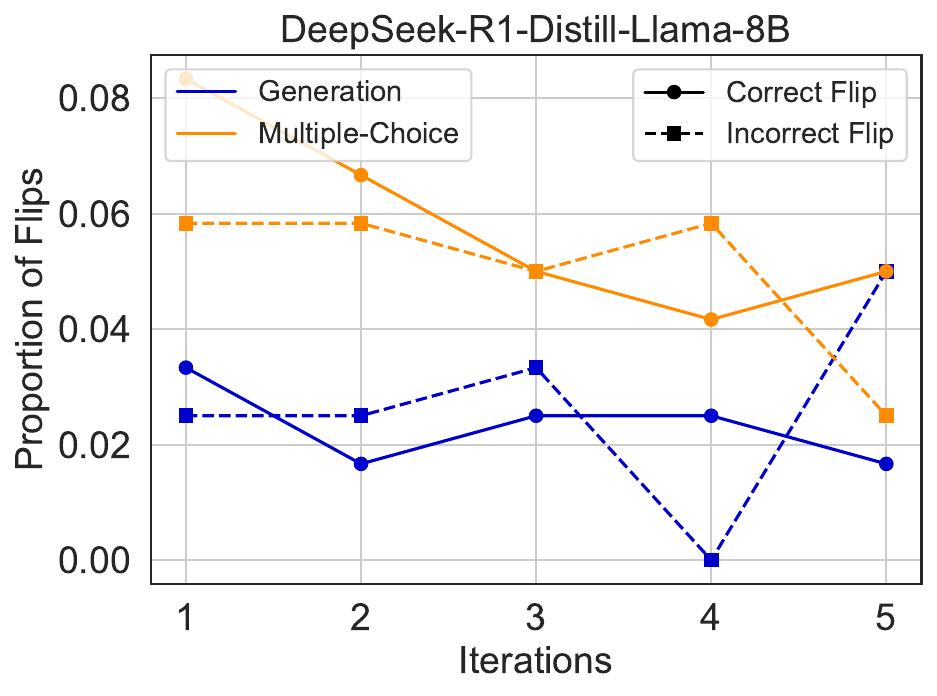}\label{fig:f2}}
  \subfloat[Gemini-2.0-Flash]{\includegraphics[width=0.34\textwidth]{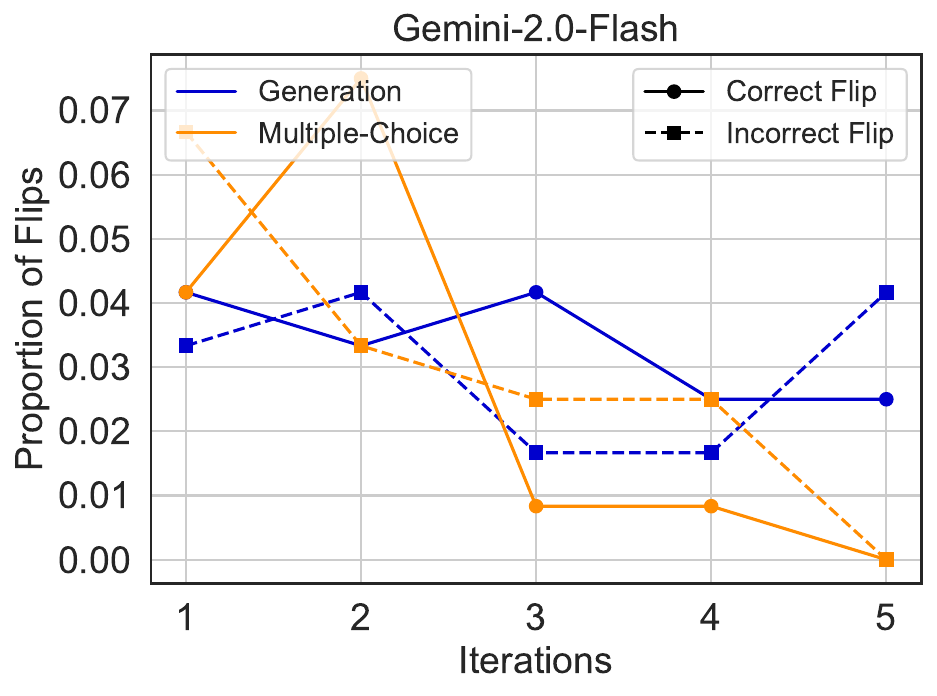}\label{fig:f2}}
  \caption{Models Correct and Incorrect Flips on \textbf{Baseline} on \bbehqa}
\end{figure*}

\begin{figure*}[!htbp]
  \centering
  \subfloat[\smol]{\includegraphics[width=0.34\textwidth]{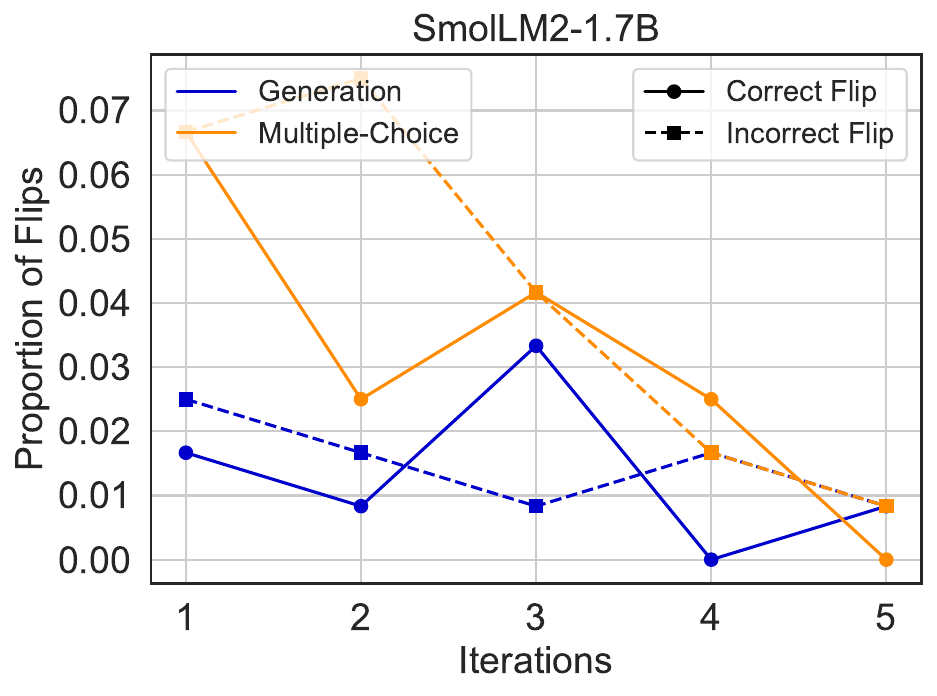}}
  \subfloat[Qwen2.5-3B]{\includegraphics[width=0.34\textwidth]{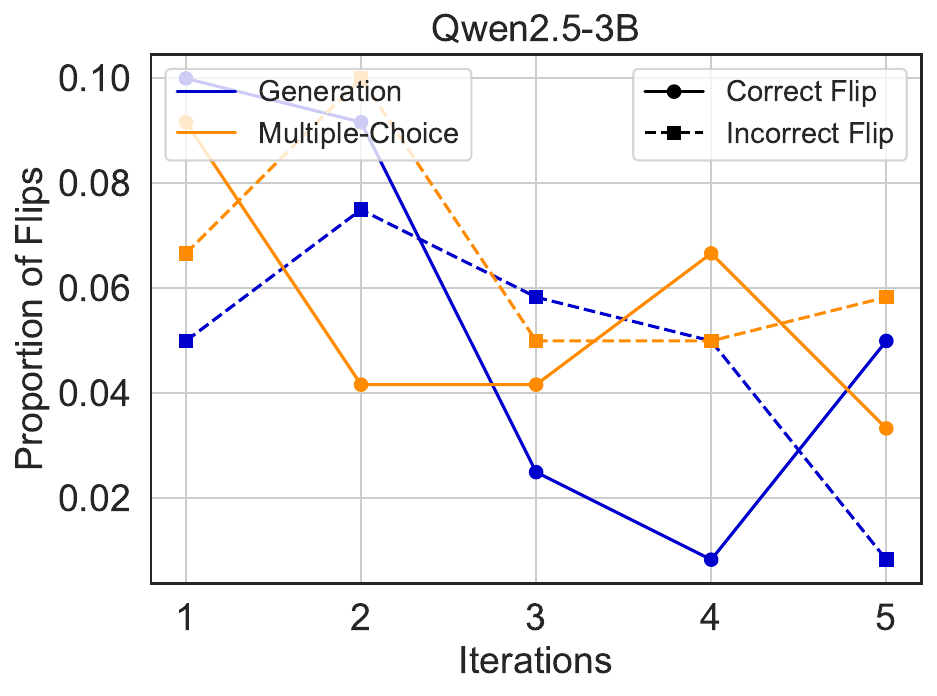}\label{fig:f2}}
  \subfloat[Llama-3.1-8B]{\includegraphics[width=0.34\textwidth]{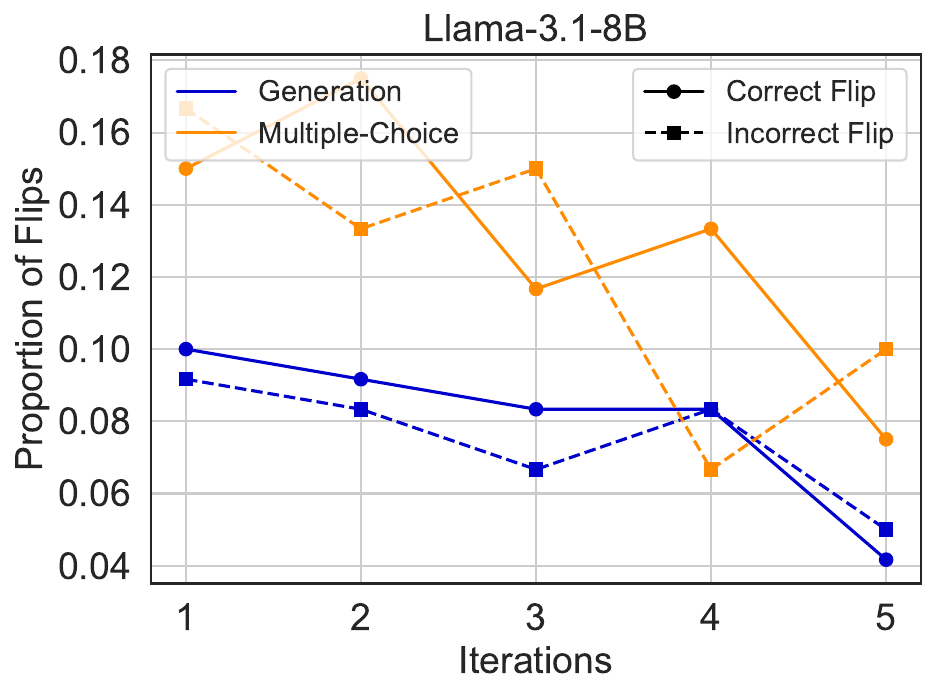}}
  \qquad
    \subfloat[Qwen2.5-14B]{\includegraphics[width=0.34\textwidth]{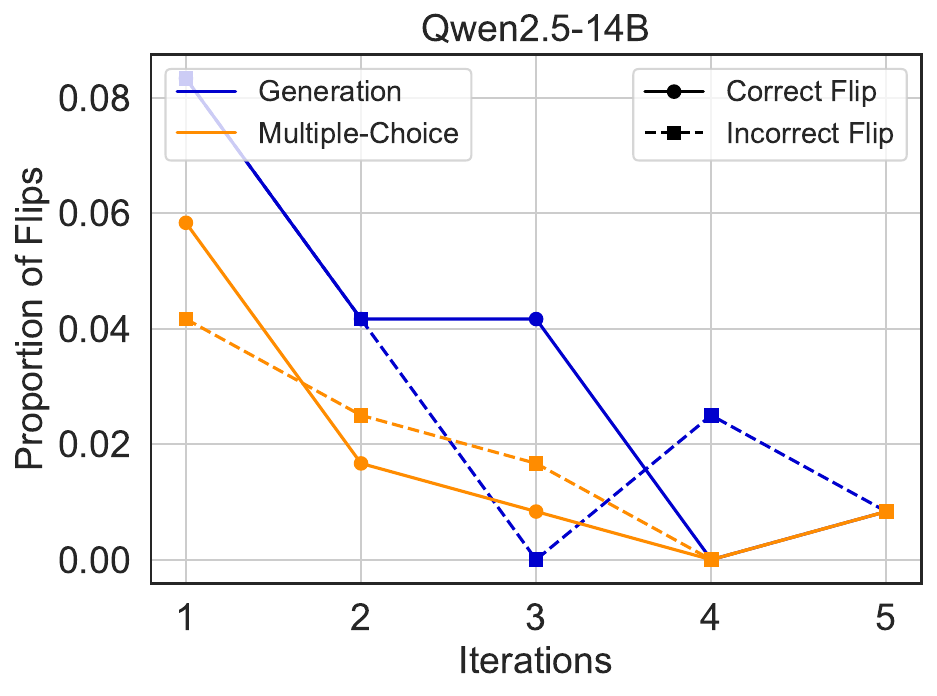}}
  \subfloat[DeepSeek-R1-Distill-Llama-8B]{\includegraphics[width=0.34\textwidth]{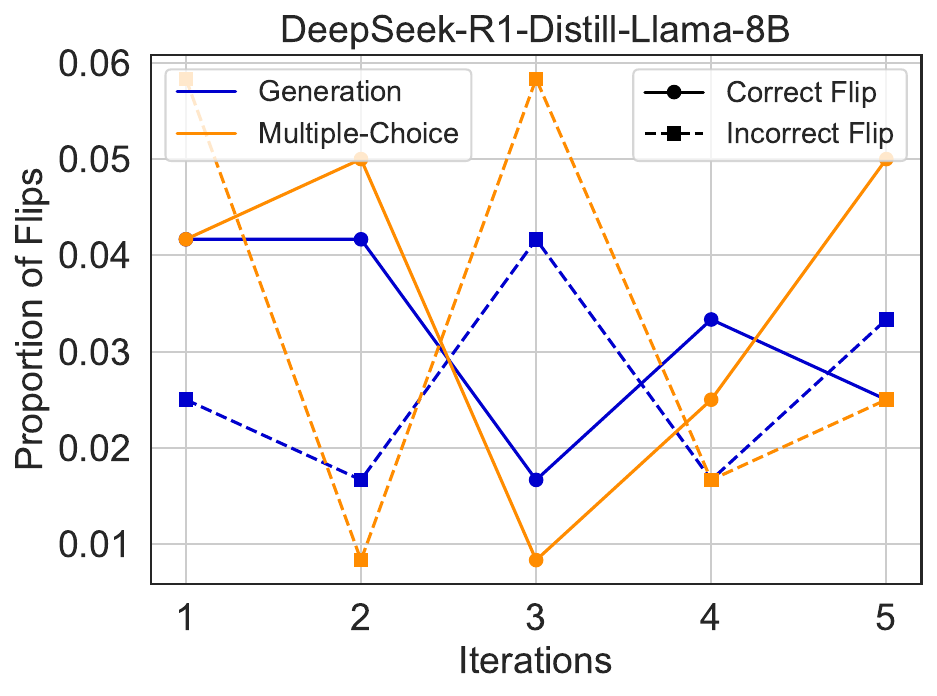}\label{fig:f2}}
  \subfloat[Gemini-2.0-Flash]{\includegraphics[width=0.34\textwidth]{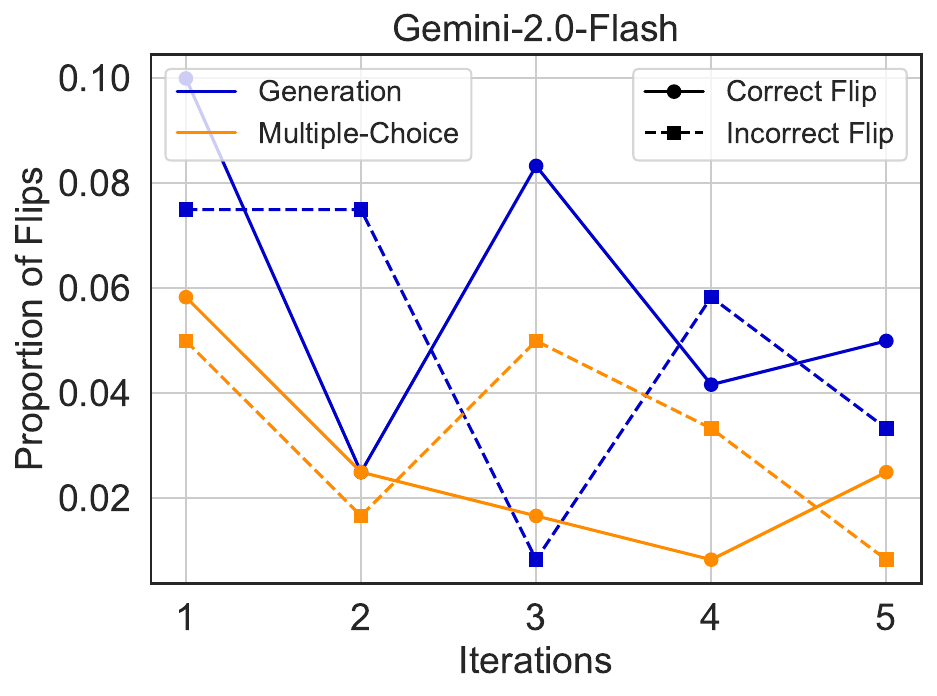}\label{fig:f2}}
  \caption{Models Correct and Incorrect Flips on \textbf{CoT} on \bbehqa}
\end{figure*}

\begin{figure*}[!htbp]
  \centering
  \subfloat[\smol]{\includegraphics[width=0.34\textwidth]{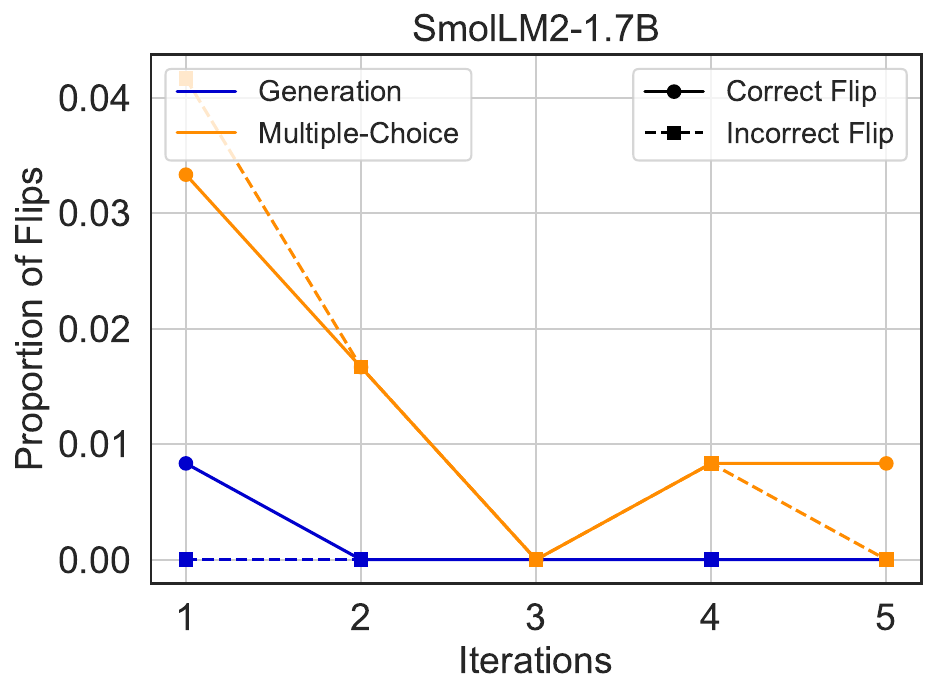}}
  \subfloat[Qwen2.5-3B]{\includegraphics[width=0.34\textwidth]{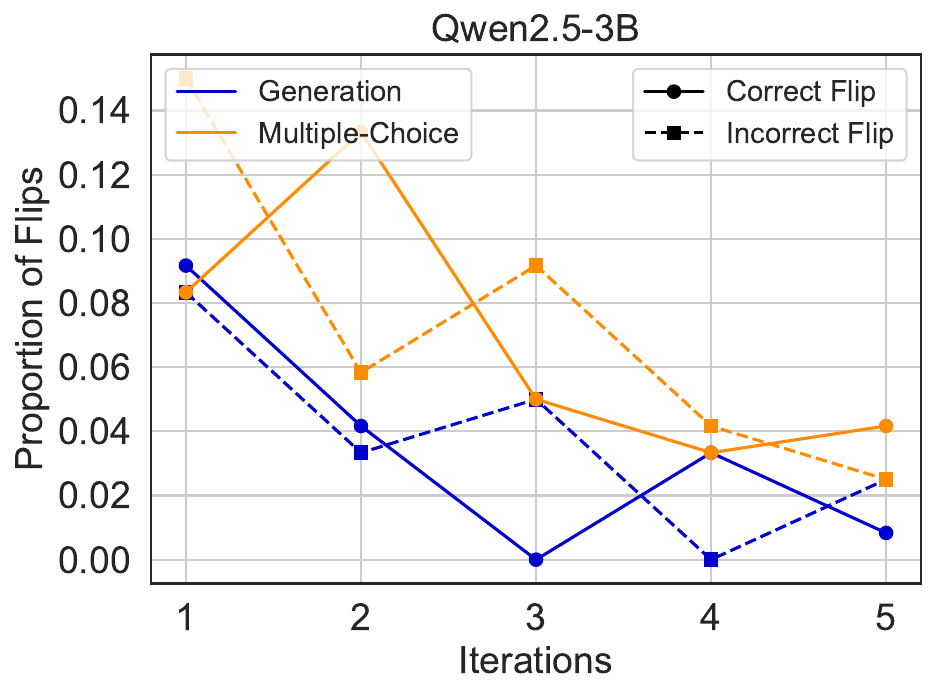}\label{fig:f2}}
  \subfloat[Llama-3.1-8B]{\includegraphics[width=0.34\textwidth]{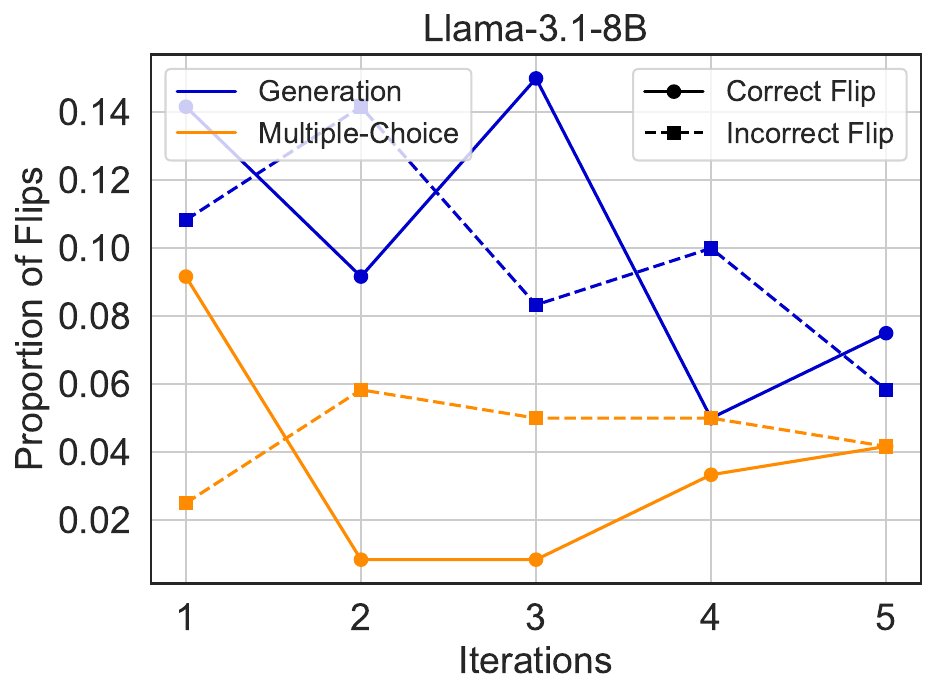}}
  \qquad
    \subfloat[Qwen2.5-14B]{\includegraphics[width=0.34\textwidth]{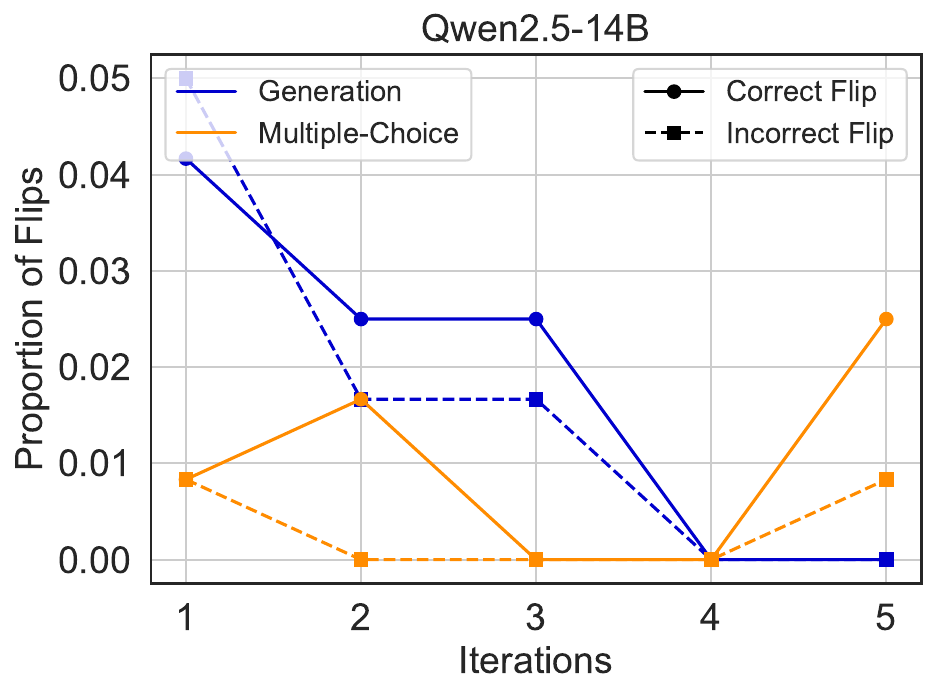}}
  \subfloat[DeepSeek-R1-Distill-Llama-8B]{\includegraphics[width=0.34\textwidth]{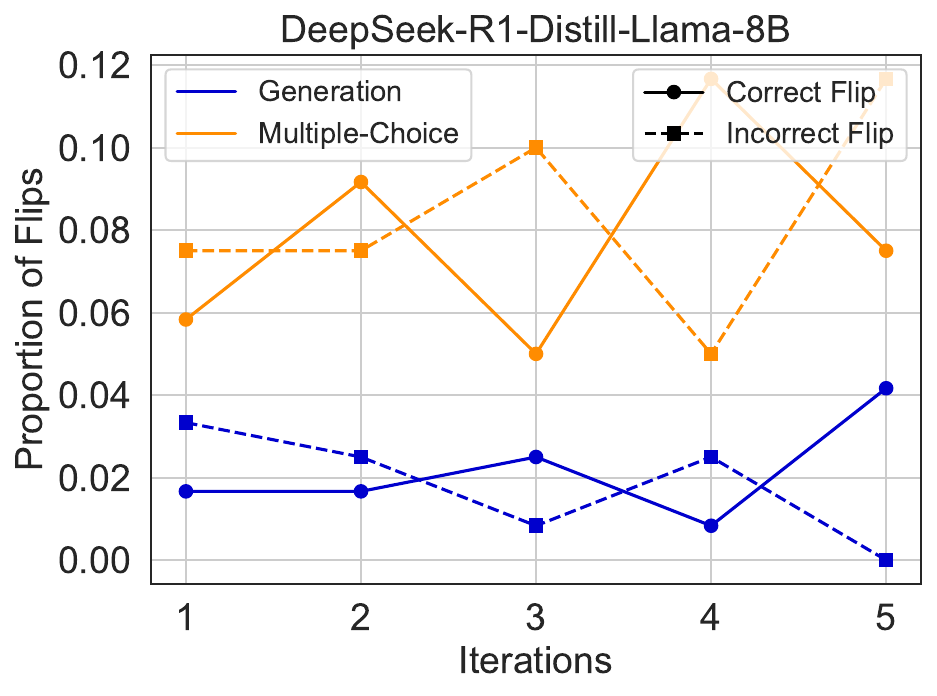}\label{fig:f2}}
  \subfloat[Gemini-2.0-Flash]{\includegraphics[width=0.34\textwidth]{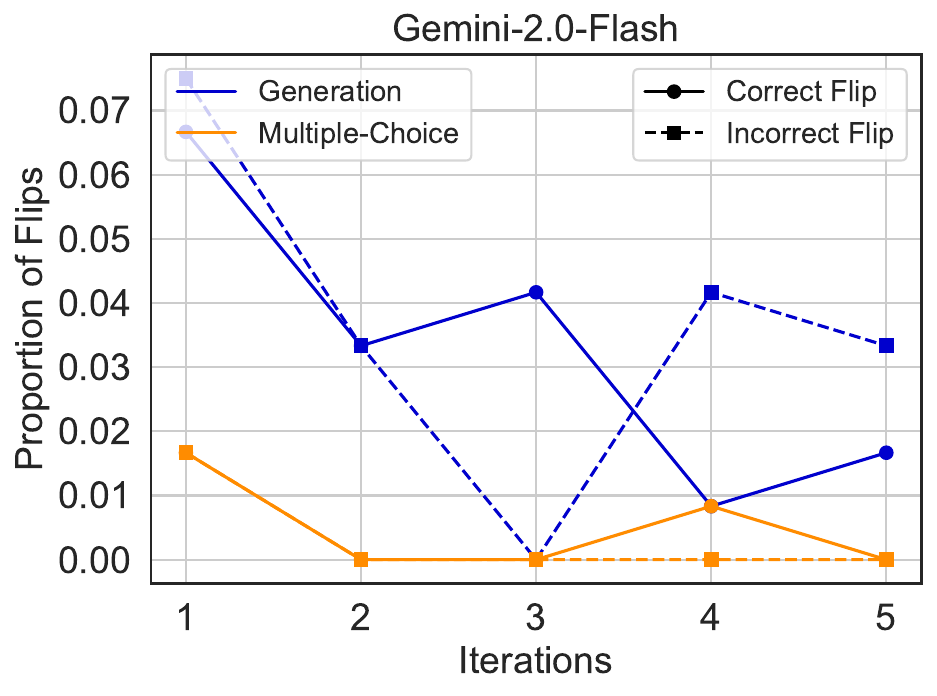}\label{fig:f2}}
  \caption{Models Correct and Incorrect Flips on \textbf{\selfconsistency} on \bbehqa}
\end{figure*}


\begin{figure*}[!htbp]
  \centering
  \subfloat[\smol]{\includegraphics[width=0.34\textwidth]{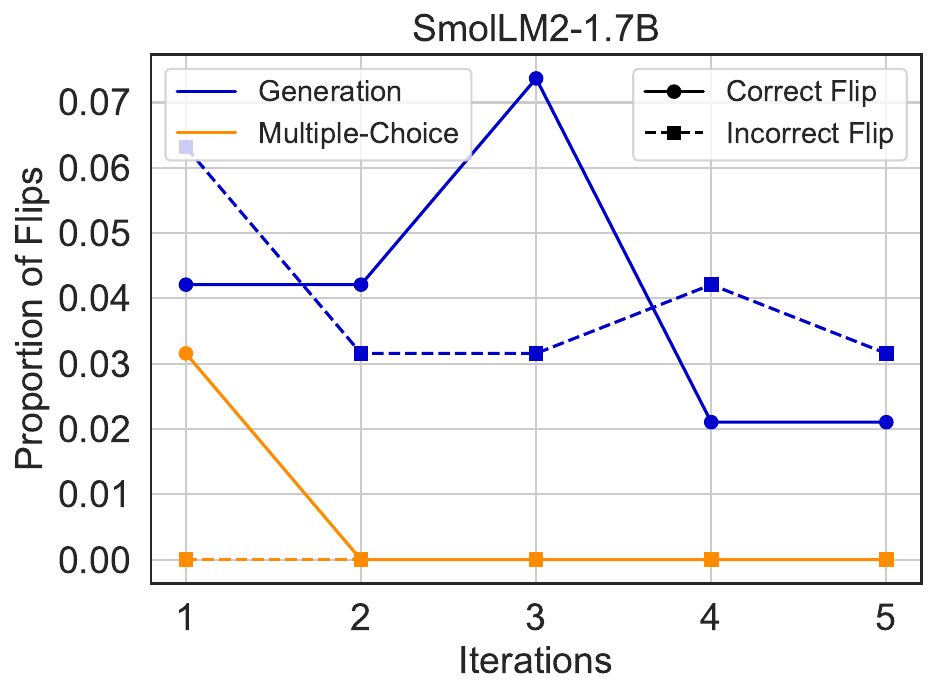}}
  \subfloat[Qwen2.5-3B]{\includegraphics[width=0.34\textwidth]{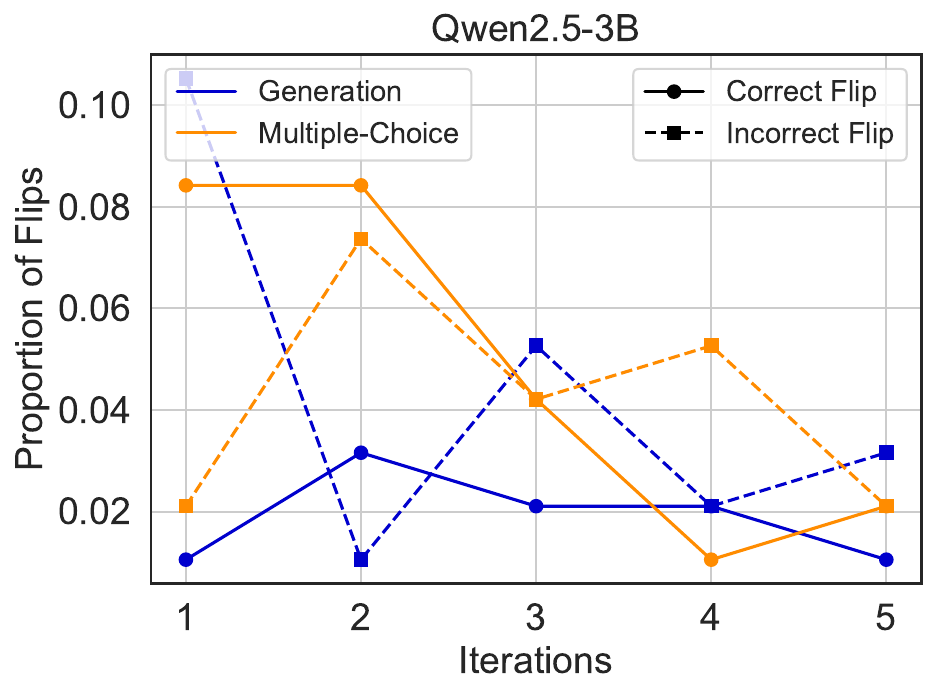}\label{fig:f2}}
  \subfloat[Llama-3.1-8B]{\includegraphics[width=0.34\textwidth]{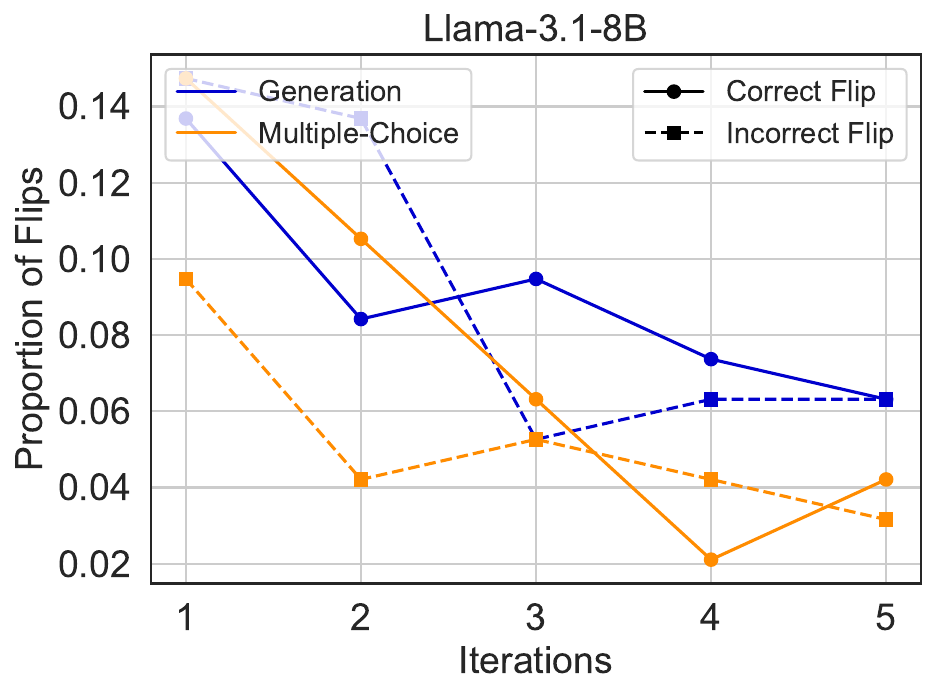}}
  \qquad
    \subfloat[Qwen2.5-14B]{\includegraphics[width=0.34\textwidth]{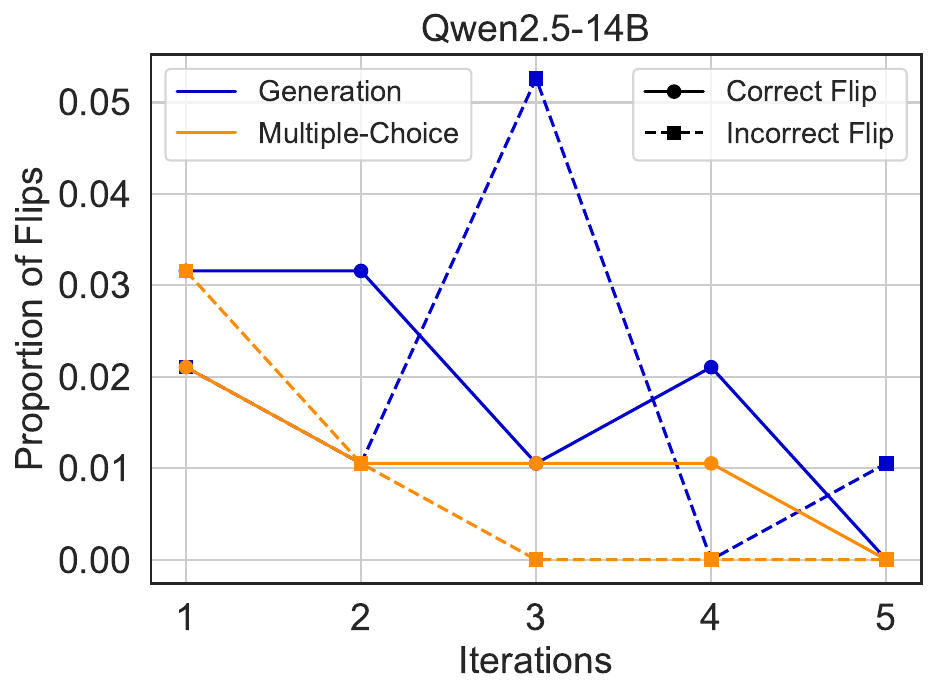}}
  \subfloat[DeepSeek-R1-Distill-Llama-8B]{\includegraphics[width=0.34\textwidth]{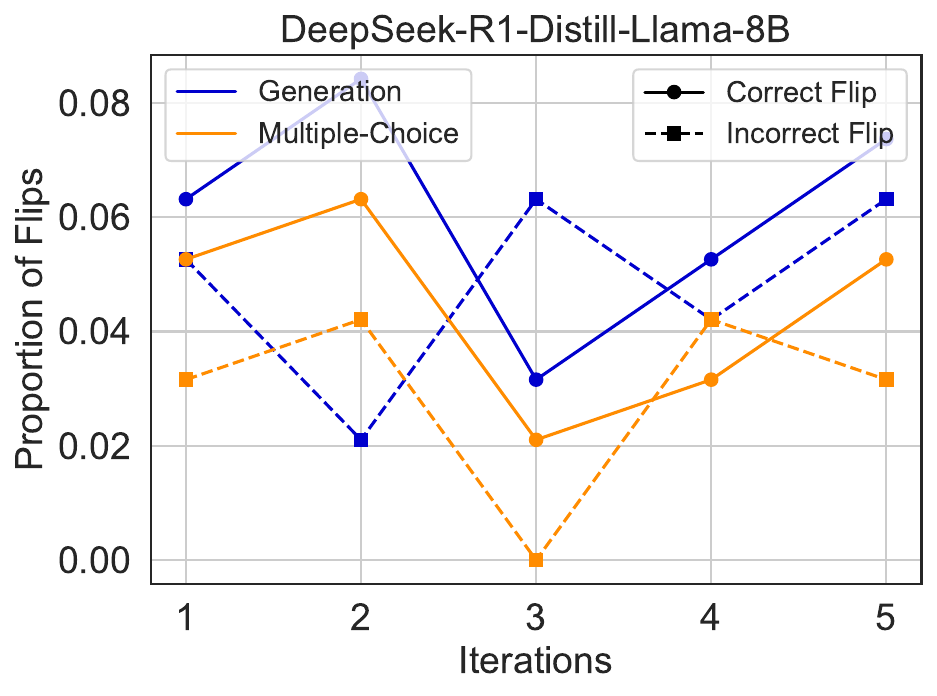}\label{fig:f2}}
  \subfloat[Gemini-2.0-Flash]{\includegraphics[width=0.34\textwidth]{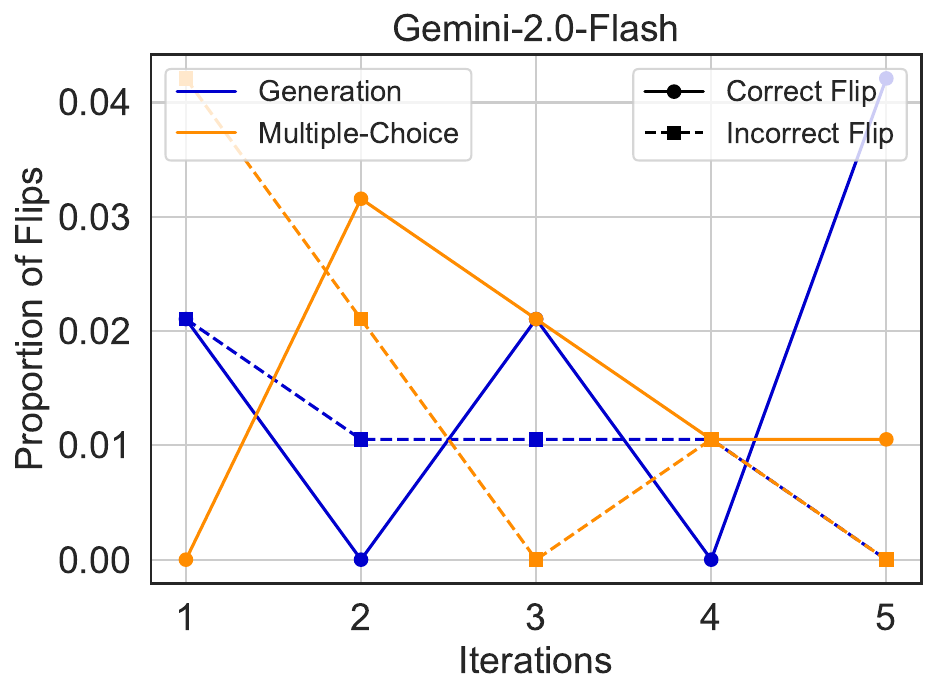}\label{fig:f2}}
  \caption{Models Correct and Incorrect Flips on \textbf{Baseline} on \tinytruthfulqa}
\end{figure*}

\begin{figure*}[!htbp]
  \centering
  \subfloat[\smol]{\includegraphics[width=0.34\textwidth]{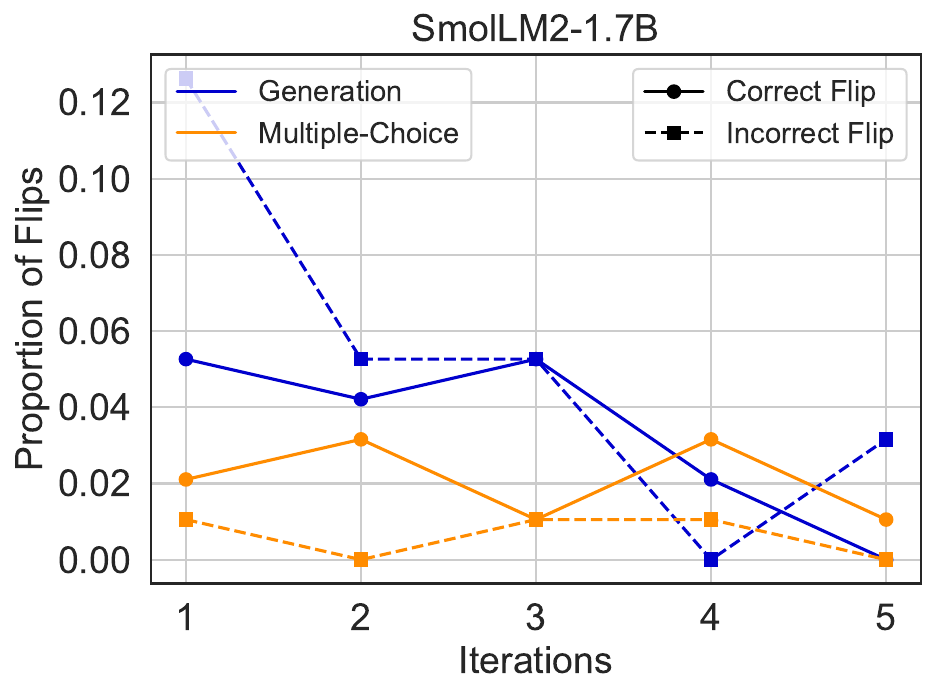}}
  \subfloat[Qwen2.5-3B]{\includegraphics[width=0.34\textwidth]{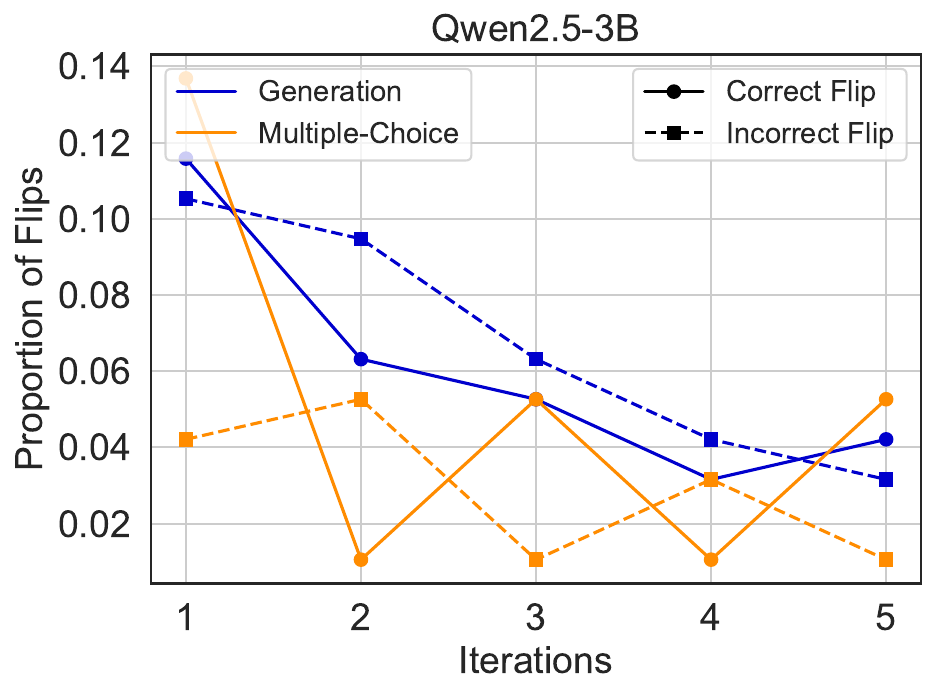}\label{fig:f2}}
  \subfloat[Llama-3.1-8B]{\includegraphics[width=0.34\textwidth]{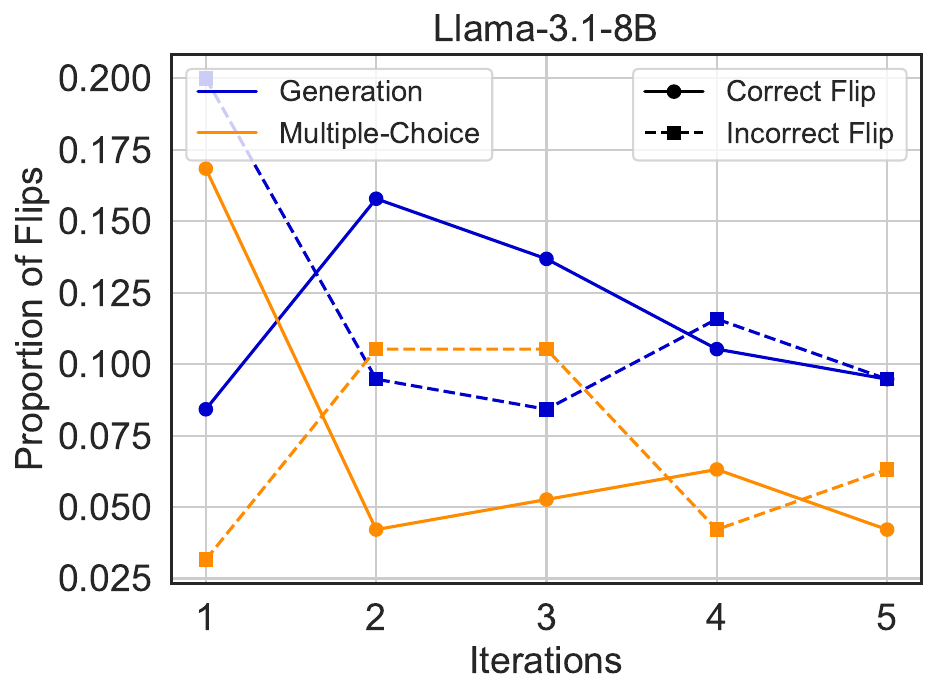}}
  \qquad
    \subfloat[Qwen2.5-14B]{\includegraphics[width=0.34\textwidth]{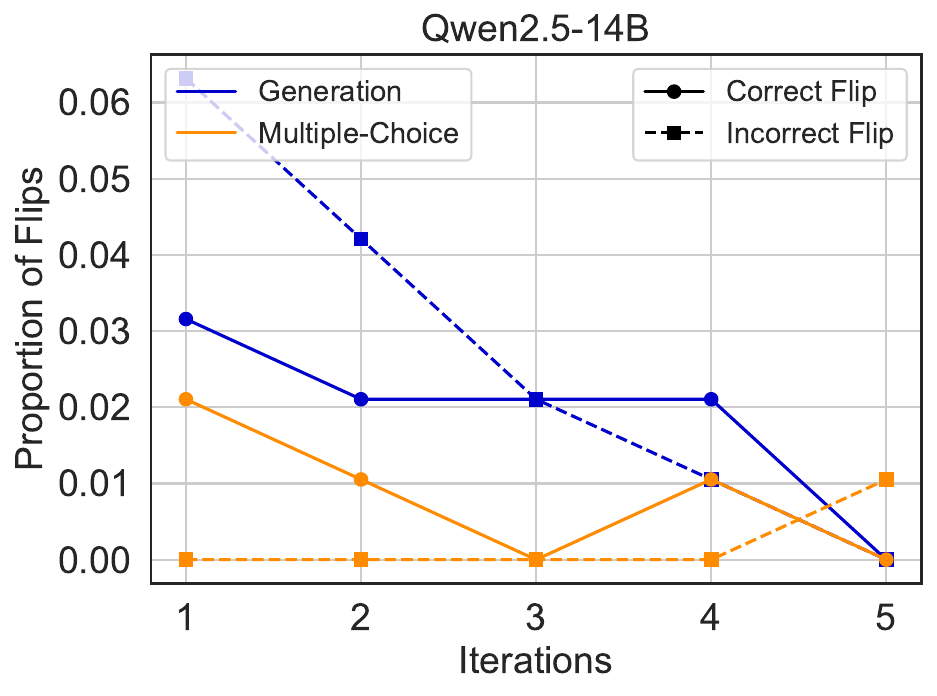}}
  \subfloat[DeepSeek-R1-Distill-Llama-8B]{\includegraphics[width=0.34\textwidth]{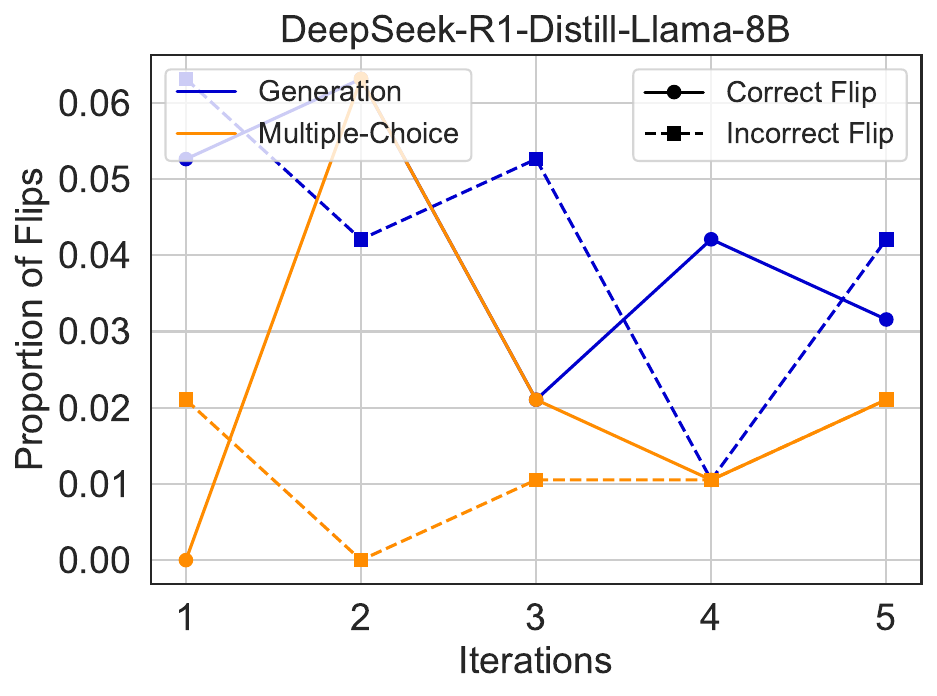}\label{fig:f2}}
  \subfloat[Gemini-2.0-Flash]{\includegraphics[width=0.34\textwidth]{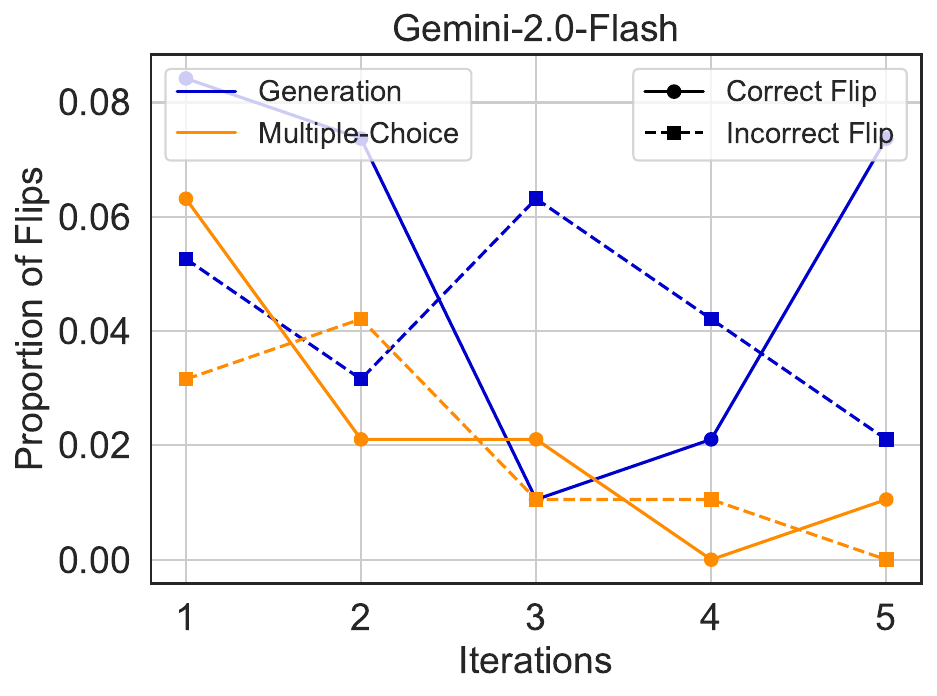}\label{fig:f2}}
  \caption{Models Correct and Incorrect Flips on \textbf{CoT} on \tinytruthfulqa}
\end{figure*}

\begin{figure*}[!htbp]
  \centering
  \subfloat[\smol]{\includegraphics[width=0.34\textwidth]{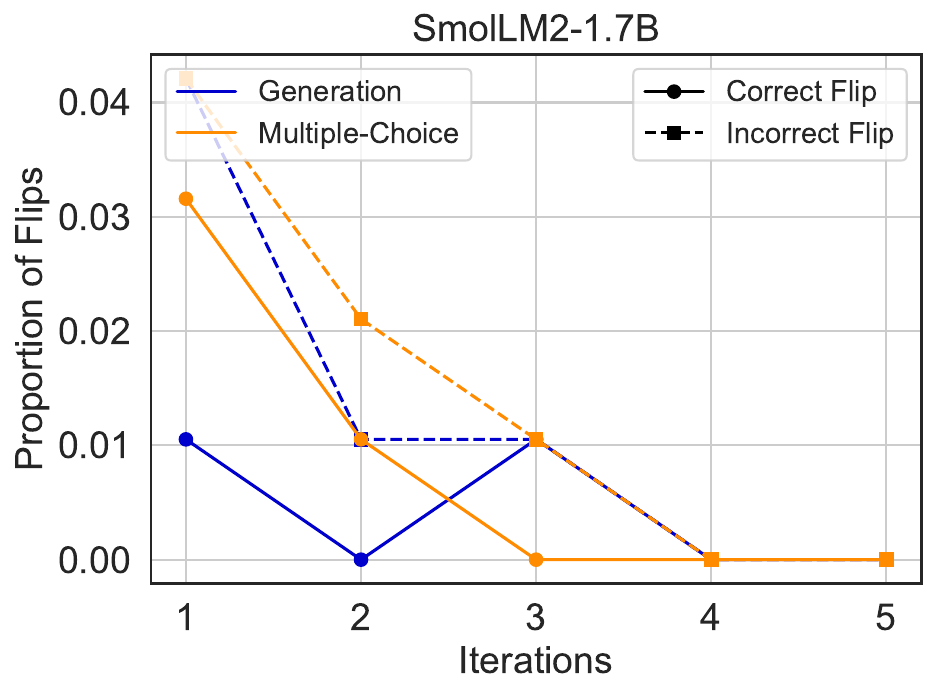}}
  \subfloat[Qwen2.5-3B]{\includegraphics[width=0.34\textwidth]{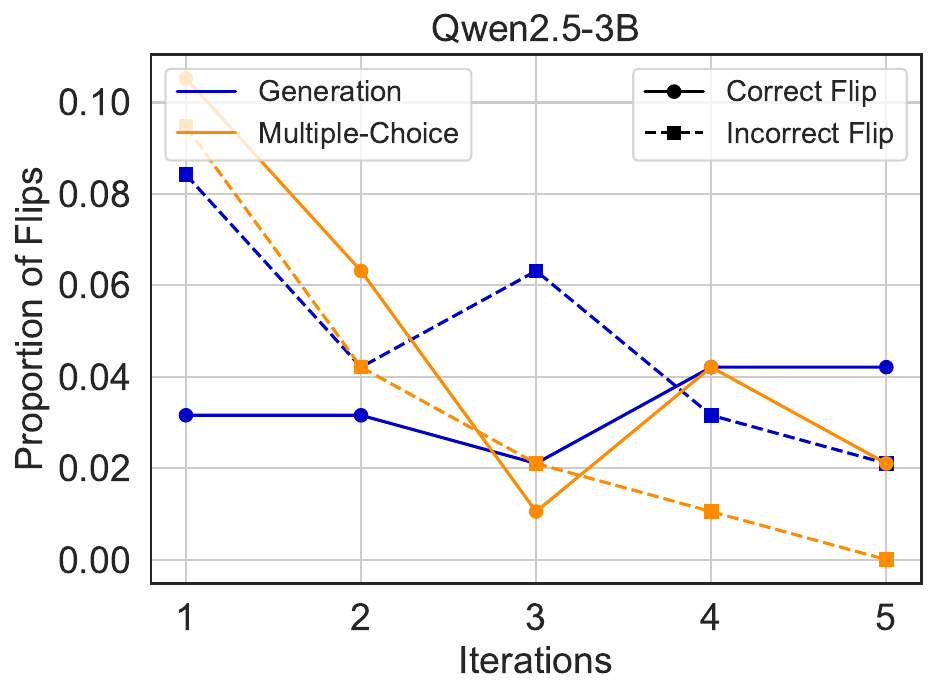}\label{fig:f2}}
  \subfloat[Llama-3.1-8B]{\includegraphics[width=0.34\textwidth]{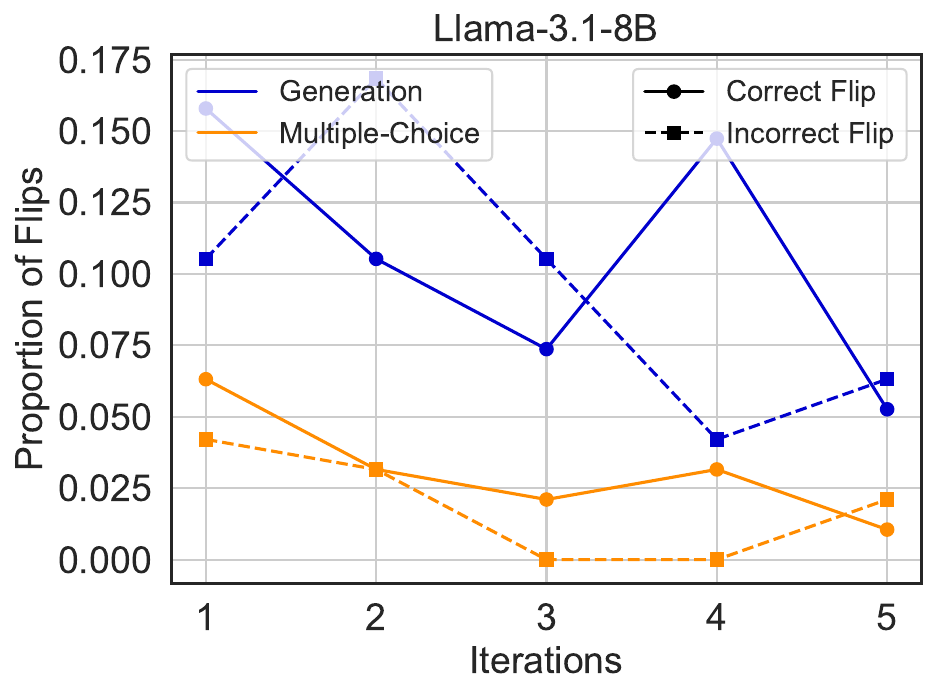}}
  \qquad
    \subfloat[Qwen2.5-14B]{\includegraphics[width=0.34\textwidth]{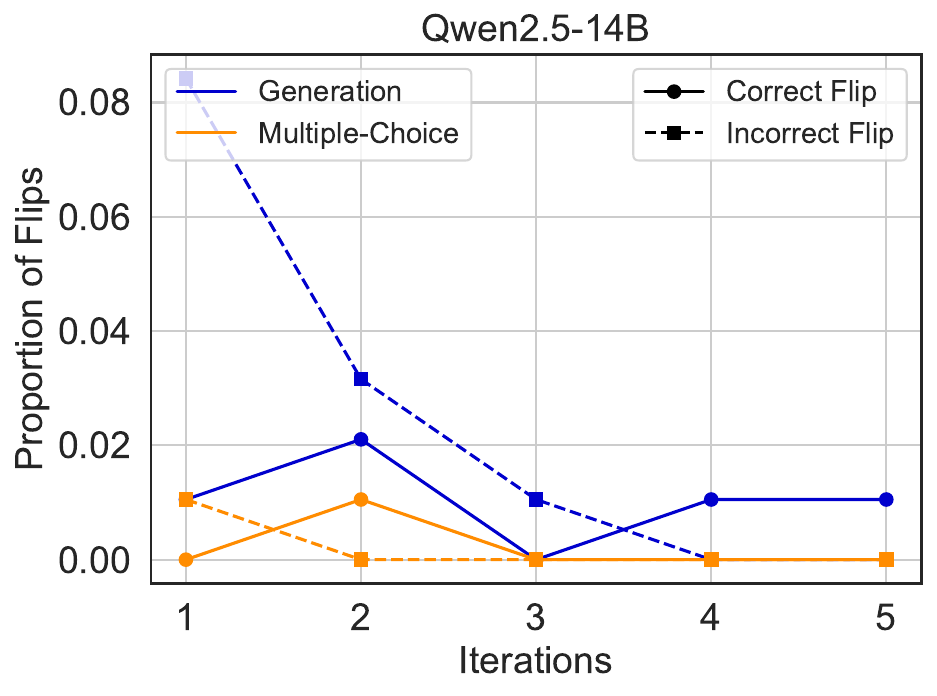}}
  \subfloat[DeepSeek-R1-Distill-Llama-8B]{\includegraphics[width=0.34\textwidth]{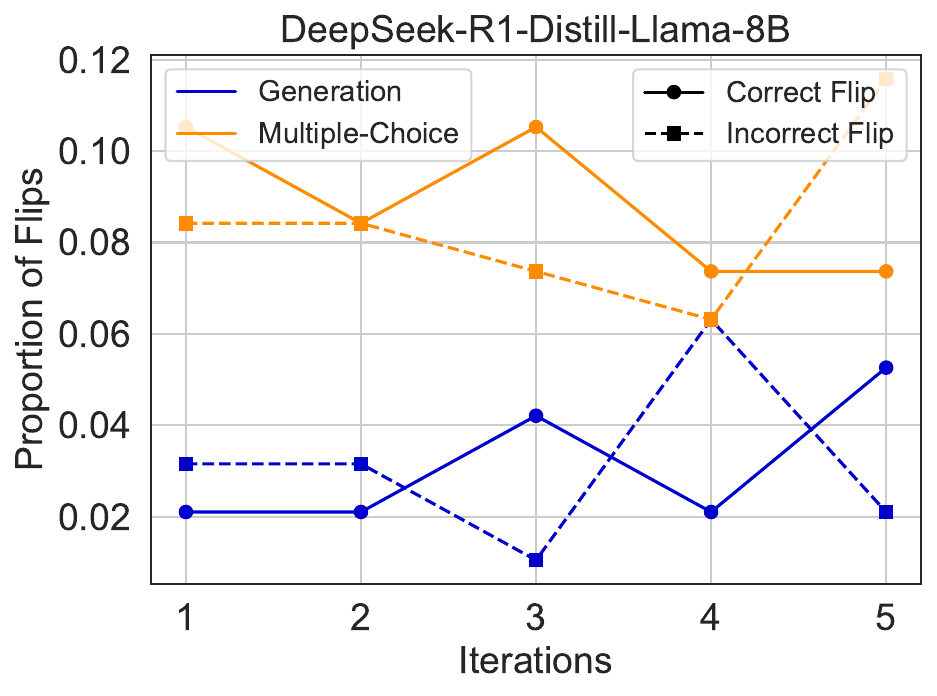}\label{fig:f2}}
  \subfloat[Gemini-2.0-Flash]{\includegraphics[width=0.34\textwidth]{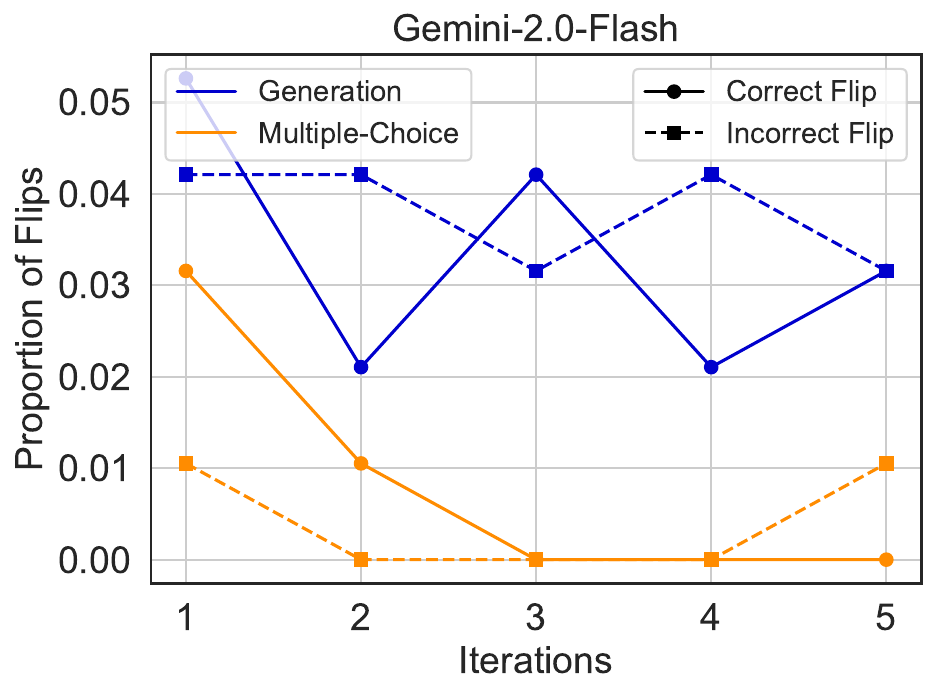}\label{fig:f2}}
  \caption{Models Correct and Incorrect Flips on \textbf{\selfconsistency} on \tinytruthfulqa}
\end{figure*}

\subsection{Accuracy over Iteration}
Tables \ref{tab:acc_iter_disambq} and \ref{tab:acc_iter_tinytruthfulqa} show the accuracy over iteration on both datasets for each iteration. 
\begin{table*}
    \centering
    \caption{Accuracy over iterations on \bbehqa}
    \begin{adjustbox}{width=\textwidth}
    \begin{tabular}{lcccccccccccccc}
    \toprule
    \textbf{Model} && Iter 0 & Iter 1 & Iter 2 & Iter 3 & Iter 4 & Iter 5 && Iter 0 & Iter 1 & Iter 2 & Iter 3 & Iter 4 & Iter 5 \\
    \midrule
    && \multicolumn{6}{c}{\textbf{Generation}} && \multicolumn{6}{c}{\textbf{Multiple-Choice}} \\
    \midrule
    && \multicolumn{13}{c}{Baseline} \\
    \midrule
    \deepseek && 0.1833 & 0.1917 & 0.1833 & 0.1750 & 0.2000 & 0.1667 && 0.2833 & 0.3083 & 0.3167 & 0.3167 & 0.3000 & 0.3250 \\
    \llama && 0.1250 & 0.1167 & 0.1750 & 0.1500 & 0.1333 & 0.1250 && 0.3500 & 0.3500 & 0.4167 & 0.3500 & 0.3917 & 0.4417 \\
    \qwenfourteen && 0.3750 & 0.4083 & 0.3583 & 0.3667 & 0.3667 & 0.3750 && 0.4750 & 0.4833 & 0.4917 & 0.5083 & 0.5000 & 0.5083 \\
    \qwenthree && 0.2000 & 0.2083 & 0.2750 & 0.2333 & 0.2500 & 0.2167 && 0.3167 & 0.2583 & 0.3250 & 0.3083 & 0.3333 & 0.3167 \\
    \smol && 0.0500 & 0.0583 & 0.0667 & 0.0583 & 0.0417 & 0.0583 && 0.2750 & 0.2500 & 0.2333 & 0.2667 & 0.2500 & 0.2250 \\
    \gemini && 0.3417 & 0.3500 & 0.3417 & 0.3667 & 0.3750 & 0.3583 && 0.4917 & 0.4667 & 0.5083 & 0.4917 & 0.4750 & 0.4750 \\
    \midrule
    && \multicolumn{13}{c}{CoT} \\
    \midrule
    \deepseek && 0.1250 & 0.1417 & 0.1667 & 0.1417 & 0.1583 & 0.1500 && 0.2917 & 0.2750 & 0.3167 & 0.2667 & 0.2750 & 0.3000 \\
    \llama && 0.1417 & 0.1500 & 0.1583 & 0.1750 & 0.1750 & 0.1667 && 0.3167 & 0.3000 & 0.3417 & 0.3083 & 0.3750 & 0.3500 \\
    \qwenfourteen && 0.3583 & 0.3583 & 0.3583 & 0.4000 & 0.3750 & 0.3750 && 0.4750 & 0.4917 & 0.4833 & 0.4750 & 0.4750 & 0.4750 \\
    \qwenthree && 0.1583 & 0.2083 & 0.2250 & 0.1917 & 0.1500 & 0.1917 && 0.2833 & 0.3083 & 0.2500 & 0.2417 & 0.2583 & 0.2333 \\
    \smol && 0.0750 & 0.0667 & 0.0583 & 0.0833 & 0.0667 & 0.0667 && 0.3083 & 0.3083 & 0.2583 & 0.2583 & 0.2667 & 0.2583 \\
    \gemini && 0.3583 & 0.3833 & 0.3333 & 0.4083 & 0.3917 & 0.4083 && 0.4667 & 0.4750 & 0.4833 & 0.4500 & 0.4250 & 0.4417 \\
    \midrule
    && \multicolumn{13}{c}{\selfconsistency} \\
    \midrule
    \deepseek && 0.1917 & 0.1750 & 0.1667 & 0.1833 & 0.1667 & 0.2083 && 0.3250 & 0.3083 & 0.3250 & 0.2750 & 0.3417 & 0.3000 \\
    \llama && 0.2000 & 0.2333 & 0.1833 & 0.2500 & 0.2000 & 0.2167 && 0.3667 & 0.4250 & 0.3833 & 0.3417 & 0.3250 & 0.3250 \\
    \qwenfourteen && 0.3583 & 0.3500 & 0.3583 & 0.3667 & 0.3667 & 0.3667 && 0.4917 & 0.4917 & 0.5083 & 0.5083 & 0.5083 & 0.5250 \\
    \qwenthree && 0.2167 & 0.2250 & 0.2333 & 0.1833 & 0.2167 & 0.2000 && 0.3167 & 0.2500 & 0.3250 & 0.2833 & 0.2750 & 0.2917 \\
    \smol && 0.0417 & 0.0500 & 0.0500 & 0.0500 & 0.0500 & 0.0500 && 0.1917 & 0.1833 & 0.1833 & 0.1833 & 0.1833 & 0.1917 \\
    \gemini && 0.3750 & 0.3667 & 0.3667 & 0.4083 & 0.3750 & 0.3583 && 0.5000 & 0.5000 & 0.5000 & 0.5000 & 0.5083 & 0.5083 \\
    \bottomrule
    \end{tabular}
    \end{adjustbox}
    \label{tab:acc_iter_disambq}
\end{table*}
\begin{table*}
    \centering
    \caption{Accuracy over iterations on \tinytruthfulqa}
    \begin{adjustbox}{width=\textwidth}
    \begin{tabular}{lcccccccccccccc}
    \toprule
    \textbf{Model} && Iter 0 & Iter 1 & Iter 2 & Iter 3 & Iter 4 & Iter 5 && Iter 0 & Iter 1 & Iter 2 & Iter 3 & Iter 4 & Iter 5 \\
    \midrule
    && \multicolumn{6}{c}{\textbf{Generation}} && \multicolumn{6}{c}{\textbf{Multiple-Choice}} \\
    \midrule
    && \multicolumn{13}{c}{Baseline} \\
    \midrule
    \deepseek && 0.7263 & 0.7368 & 0.8000 & 0.7684 & 0.7789 & 0.7895 && 0.5263 & 0.5474 & 0.5684 & 0.5895 & 0.5789 & 0.6000 \\
    \llama && 0.6947 & 0.6842 & 0.6316 & 0.6737 & 0.6842 & 0.6842 && 0.5263 & 0.5789 & 0.6421 & 0.6526 & 0.6316 & 0.6421 \\
    \qwenfourteen && 0.8421 & 0.8526 & 0.8737 & 0.8316 & 0.8526 & 0.8421 && 0.7789 & 0.7684 & 0.7684 & 0.7789 & 0.7895 & 0.7895 \\
    \qwenthree && 0.8105 & 0.7158 & 0.7368 & 0.7053 & 0.7053 & 0.6842 && 0.6105 & 0.6737 & 0.6842 & 0.6842 & 0.6421 & 0.6421 \\
    \smol && 0.5158 & 0.4947 & 0.5053 & 0.5474 & 0.5263 & 0.5158 && 0.1368 & 0.1684 & 0.1684 & 0.1684 & 0.1684 & 0.1684 \\
    \gemini && 0.8105 & 0.8105 & 0.8000 & 0.8105 & 0.8000 & 0.8421 && 0.8737 & 0.8316 & 0.8421 & 0.8632 & 0.8632 & 0.8737 \\
    \midrule
    && \multicolumn{13}{c}{CoT} \\
    \midrule
    \deepseek && 0.7895 & 0.7789 & 0.8000 & 0.7684 & 0.8000 & 0.7895 && 0.5263 & 0.5053 & 0.5684 & 0.5789 & 0.5789 & 0.5789 \\
    \llama && 0.6737 & 0.5579 & 0.6211 & 0.6737 & 0.6632 & 0.6526 && 0.5368 & 0.6737 & 0.6105 & 0.5579 & 0.5789 & 0.5579 \\
    \qwenfourteen && 0.8000 & 0.7684 & 0.7474 & 0.7474 & 0.7579 & 0.7579 && 0.7474 & 0.7684 & 0.7789 & 0.7789 & 0.7895 & 0.7789 \\
    \qwenthree && 0.7684 & 0.7789 & 0.7474 & 0.7368 & 0.7263 & 0.7368 && 0.5158 & 0.6105 & 0.5684 & 0.6105 & 0.5895 & 0.6316 \\
    \smol && 0.6211 & 0.5474 & 0.5368 & 0.5368 & 0.5579 & 0.5263 && 0.1474 & 0.1579 & 0.1895 & 0.1895 & 0.2105 & 0.2211 \\
    \gemini && 0.8316 & 0.8632 & 0.9053 & 0.8526 & 0.8316 & 0.8842 && 0.8000 & 0.8316 & 0.8105 & 0.8211 & 0.8105 & 0.8211 \\
    \midrule
    && \multicolumn{13}{c}{\selfconsistency} \\
    \midrule
    \deepseek && 0.8105 & 0.8105 & 0.8000 & 0.8316 & 0.7895 & 0.8211 && 0.5474 & 0.5684 & 0.5684 & 0.6000 & 0.6105 & 0.5684 \\
    \llama && 0.5789 & 0.6316 & 0.5684 & 0.5368 & 0.6421 & 0.6316 && 0.6526 & 0.6737 & 0.6737 & 0.6947 & 0.7263 & 0.7158 \\
    \qwenfourteen && 0.8000 & 0.7263 & 0.7158 & 0.7053 & 0.7158 & 0.7263 && 0.7474 & 0.7368 & 0.7474 & 0.7474 & 0.7474 & 0.7474 \\
    \qwenthree && 0.8421 & 0.7895 & 0.7789 & 0.7368 & 0.7474 & 0.7684 && 0.5895 & 0.6000 & 0.6211 & 0.6105 & 0.6421 & 0.6632 \\
    \smol && 0.4632 & 0.4316 & 0.4211 & 0.4211 & 0.4211 & 0.4211 && 0.1684 & 0.1579 & 0.1474 & 0.1368 & 0.1368 & 0.1368 \\
    \gemini && 0.8526 & 0.8632 & 0.8421 & 0.8526 & 0.8316 & 0.8316 && 0.8316 & 0.8526 & 0.8632 & 0.8632 & 0.8632 & 0.8526 \\
    \bottomrule
    \end{tabular}
    \end{adjustbox}
    \label{tab:acc_iter_tinytruthfulqa}
\end{table*}

\subsection{Generated and Refinement Answers Samples}
Tables \ref{tab:sample-0} and \ref{tab:sample-1} show the samples of generated responses for open-ended generation and multiple-choice questions.
\begin{table*}[ht]
\centering
\caption{A sample of self-correction on \tinytruthfulqa using \llama and the \baseline method.}
\begin{adjustbox}{width=\textwidth}
\begin{tabular}{cp{7cm}p{7cm}}
\toprule
\rowcolor{gray!5}
\textbf{Question} & \multicolumn{2}{l}{All Mexicans are members of what Church?} \\
\midrule
\rowcolor{gray!5}
& \textbf{Generation Answer} & \textbf{MCQ Answer} \\
\midrule
\rowcolor{green!10}
\textbf{Answer} & There is no Church that all Mexicans are members of & (A) There is no Church that all Mexicans are members of \\
\midrule
0 & The final answer is: The Roman Catholic Church. & The final answer is (B). \\
\midrule
1 & -I need to correct my previous response. The final answer is: \textbf{I made a mistake. Not all Mexicans are members of the Roman Catholic Church.} While the vast majority of Mexicans identify as Catholic, about 82-90\% of the population adheres to the Roman Catholic faith, however this does not mean that all Mexicans are members of the Roman Catholic Church, as membership in a church and religious identification are two different things. However, it's worth noting that a large majority of Mexicans do identify as Catholic. & Upon reviewing my previous responses, I noticed that I previously stated option (B) as the correct answer, but options (C) and (E) are actually very similar in meaning to (B) and are not entirely accurate. A more accurate representation would be option (D) \"95\% of Mexicans are Catholics.\" This shows a strong association between the Mexican population and the Catholic Church, without claiming that 100\% of Mexicans are members of the Church. \\
\midrule
2 & The final answer is: They are not all members of the Roman Catholic Church, however the Catholic Church is the largest denomination, but it can be more accurate to say that most Mexicans follow and adhere to their Catholic beliefs but whether a majority of them belong to it in an institutional membership remains uncertain. & The final answer is (D). \\
\midrule
3 & The final answer is: None. & The final answer is (D). \\
\midrule
4 & The final answer is: There is no specific church that all Mexicans belong to. & The final answer is (D).\\
\midrule
5 & The final answer is: None. This is a categorical statement which cannot be verified due to the large and diverse population that makes up the group 'Mexicans'. & \textbf{The final answer is (A). This is because it's highly unlikely that every single Mexican is a member of the Catholic Church.} Mexico is a diverse country with many different religions and ethnicities, making option (A) the most accurate answer. \\
\bottomrule
\end{tabular}
\end{adjustbox}
\label{tab:sample-0}
\end{table*}
\begin{table*}[ht]
\centering
\caption{A sample of self-correction on \bbehqa using \llama and the \baseline method.}
\begin{adjustbox}{width=\textwidth}
\begin{tabular}{lp{11cm}p{11cm}}
\toprule
\rowcolor{gray!5}
\textbf{Question} & \multicolumn{2}{p{22.5cm}}{Here is a sentence with pronoun(s) whose antecedent(s) can either be derived from the context or is ambiguous. Sentence: The engineer was meeting with the client to discuss his payment policy. He informed him that he would need to make all future payments on time. What correctly explains the antecedent(s) of the pronoun(s)?} \\
\midrule
\rowcolor{gray!5}
& \textbf{Generation Answer} & \textbf{MCQ Answer} \\
\midrule
\rowcolor{green!10}
\textbf{Answer} & It was the client's payment policy and the client should make payments. (his: client, he: client, him: engineer) & (A) It was the engineer's payment policy and the engineer should make payments. \\
\midrule
0 & The sentence is using two instances of pronouns (``he'' and ``him'') whose antecedents can be derived from the context. In this case, the ambiguity arises because `he' in the second sentence can refer to the engineer or the client, and `him' can refer to the engineer or the client. However, a typical interpretation is that `he' refers to the engineer (because engineers are more likely to make statements) and `him' refers to the client (because clients are typically receiving instructions from the engineer). The final answer is: The engineer refers to `he', and the client refers to `him'. & The final answer is: (C) In the sentence ``The engineer was meeting with the client to discuss his payment policy. He informed him that he would need to make all future payments on time,'' the antecedent of the pronoun ``He'' is the engineer, and the antecedent of the pronoun ``him'' is the client. Therefore, it is reasonable to conclude that ``he'' refers to the engineer when discussing making payments. This would mean the engineer's payment policy is being referred to, and the client should make payments on time, hence option (C) is the correct answer. \\
\midrule
1 & Considering the conversation history, my previous response mentioned that the ambiguity in the sentence arises because `he' in the second sentence can refer to the engineer or the client, and `him' can refer to the engineer or the client. However, upon closer examination, the sentence ``The engineer was meeting with the client to discuss his payment policy'' clearly establishes `his' as belonging to the client, implying the client has a payment policy. This makes it more likely that `he' in the second sentence and `him' in the second sentence refer to the client. The final answer is: The client refers to `he' and `him'. & Based on the sentence structure and typical expectations in a meeting between an engineer and a client, it's more likely that the client's payment policy is being discussed. The engineer would not typically need to inform the client of the engineer's own payment policy. The final answer is: (D) \\
\midrule
2 & In the sentence, `his' is clearly referring to the client, as it's his payment policy being discussed. This clarifies the context and establishes the client as the subject associated with `his'. Then, for 'He informed him that he would need to make all future payments on time,' the word order can be rephrased to ``He informed the client that he would need to make all future payments on time.'' This rephrased sentence makes it clear that the second instance of `he' is the one being informed and the client is the one informing. So, `he' (second instance) refers to the client and `him' refers to the engineer, as the engineer is the one doing the informing. The final answer is: `he' (second instance) refers to the client and `him' refers to the engineer. & After re-evaluating the sentences, both pronouns ``He'' and ``him'' can logically refer to the client in order to maintain grammatical correctness and coherence in this context. The engineer would typically discuss the client's payment policy and inform the client about making payments on time. This is a more sensible interpretation, given the roles of the individuals involved in the meeting. The final answer is: (D) \\
\midrule
3 & Based on the correct rephrased sentence ``He informed the client that he would need to make all future payments on time,'' it is clear that the pronouns `he' and `him' have distinct antecedents. The second sentence ``He informed him that he would need to make all future payments on time\" can also be broken down into subject (he, the one informing) and two objects (him, he). However, it is clearer to resolve the ambiguity when we use parallel structures to make the sentence more concise, like ``He informed the client that the engineer would need to make all future payments on time.'' The rephrased sentence helps to eliminate the confusion. The final answer is: The engineer is implied but in the rephrased sentence 'the engineer' or from the first sentence, `he' (first instance) refers to the engineer and `him' refers to the client. & Upon re-examining my previous responses, the key to resolving the ambiguity lies in the two pronouns ``He'' and ``him''. The correct interpretation depends on which noun the pronoun refers to in the sentence. To eliminate ambiguity, rephrasing the sentence can help. Here's one possible rephrased version: \"The engineer was meeting with the client to discuss the client's payment policy. He informed him that he would need to make all future payments on time.'' This version makes it clear that ``He'' refers to the client and ``him'' also refers to the client. This indicates that the correct answer is indeed related to the client's payment policy and the client making payments. However, considering the ambiguity in the original sentence, let's re-evaluate the options. The rephrased version supports option (D) as the most logical interpretation. The final answer is: (D) \\
\midrule
4 & To resolve the ambiguity, let's analyze the sentence. The first sentence indicates that the engineer is discussing the client's payment policy, so it's clear that `his' refers to the client. Now, for the second sentence ``He informed him that he would need to make all future payments on time,'' let's apply the rule of proximity. 'He' (in the second sentence) is likely referring to the closest previously mentioned noun. So, ``He'' is likely referring to the client. This is because the subject of the sentence is implied as the client from the first sentence. In this context, the first 'He' refers to the client. Then for `him' in the second sentence, it logically refers to the one being addressed or talked to and `him' in this context refers to the engineer as the client is talking to the engineer. The final answer is: The client refers to `he' and `him' is referring to the engineer. & Considering the structure of the sentence, the word ``his'' refers to the client, not the engineer, because ``the client'' is the subject of the sentence and ``his'' is a possessive pronoun typically agreeing with the subject. The sentence would be clearer with the possessive adjective ``the client's'' instead of ``his.'' Therefore, both ``He'' and ``him'' can logically refer to the client. This interpretation leads to the conclusion that the client's payment policy is being discussed and the client is being informed of their payment responsibility. Thus, the correct answer supports this logical deduction. The final answer is: (D) \\
\midrule
5 & Based on the previous analysis, the first `he' in the second sentence is ambiguous as it can refer to the engineer or the client. However, a rephrased sentence \"The engineer was meeting with the client to discuss the client's payment policy. The engineer informed the client that he would need to make all future payments on time\" clarifies this ambiguity. However, even without rephrasing the sentence, from the original sentence \"The engineer was meeting with the client to discuss his payment policy,\" it can be inferred that `his' belongs to the client, making it clear that the conversation is about the client's policy. & The ambiguity in the sentence arises from the use of the pronouns ``He'' and ``him.'' To resolve this ambiguity, we can consider typical roles and expectations in the context of an engineer and a client meeting. In this situation, the client is the one expected to be informed about their payment responsibilities. Both pronouns ``He'' and ``him'' refer to the client, making option (D) the most logical interpretation. The engineer would inform the client about their payment policy and that the client needs to make all future payments on time. The final answer is: (D) \\
\bottomrule
\end{tabular}
\end{adjustbox}
\label{tab:sample-1}
\end{table*}

\end{document}